\documentclass[conference]{IEEEtran}
\IEEEoverridecommandlockouts
\usepackage{cite}
\usepackage{amsmath,amssymb,amsfonts}
\usepackage{algorithm}
\usepackage{algorithmic}
\usepackage{multirow}
\usepackage{subfigure}
\usepackage{graphicx}
\usepackage{textcomp}
\usepackage{xcolor}

\def\BibTeX{{\rm B\kern-.05em{\sc i\kern-.025em b}\kern-.08em
    T\kern-.1667em\lower.7ex\hbox{E}\kern-.125emX}}
\begin{document}

\title{Meta Pattern Concern Score: A Novel Evaluation Measure with Human Values for Multi-classifiers}

\author{\IEEEauthorblockN{Yanyun Wang}
\IEEEauthorblockA{\textit{Department of Computer Science} \\
\textit{The University of Hong Kong}\\
Hong Kong, China \\
yynwang@connect.hku.hk}
\and
\IEEEauthorblockN{Dehui Du$^{*}$}
\IEEEauthorblockA{\textit{Software Engineering Institute} \\
\textit{East China Normal University}\\
Shanghai, China \\
dhdu@sei.ecnu.edu.cn}
\and
\IEEEauthorblockN{Yuanhao Liu}
\IEEEauthorblockA{\textit{Software Engineering Institute} \\
\textit{East China Normal University}\\
Shanghai, China \\
51215902094@stu.ecnu.edu.cn}
}

\maketitle

\begin{abstract}
While advanced classifiers have been increasingly used in real-world safety-critical applications, how to properly evaluate the black-box models given specific human values remains a concern in the community. Such human values include punishing error cases of different severity in varying degrees and making compromises in general performance to reduce specific dangerous cases. In this paper, we propose a novel evaluation measure named Meta Pattern Concern Score based on the abstract representation of probabilistic prediction and the adjustable threshold for the concession in prediction confidence, to introduce the human values into multi-classifiers. Technically, we learn from the advantages and disadvantages of two kinds of common metrics, namely the confusion matrix-based evaluation measures and the loss values, so that our measure is effective as them even under general tasks, and the \textit{cross entropy} loss becomes a special case of our measure in the limit. Besides, our measure can also be used to refine the model training by dynamically adjusting the learning rate. The experiments on four kinds of models and six datasets confirm the effectiveness and efficiency of our measure. And a case study shows it can not only find the ideal model reducing 0.53\% of dangerous cases by only sacrificing 0.04\% of training accuracy, but also refine the learning rate to train a new model averagely outperforming the original one with a 1.62\% lower value of itself and 0.36\% fewer number of dangerous cases.
\end{abstract}


\section{Introduction}
A classifier is a learned model that approximates a specific target function to sort given input into predicted output \cite{mitchell1997machine}. Through processing labeled data, supervised algorithms enable advanced classifiers to learn the key features needed for constructing accurate predictions automatically \cite{chollet2021deep}. While such simplicity lays a solid foundation for the widespread implementation of classifiers nowadays, it also makes the model learning a black-box process that is hard to be flexibly and properly evaluated. 

Currently, there are two types of metrics commonly used in practice to assess the quality of learning-based classifiers. One is the \textbf{evaluation measure} based on confusion matrix \cite{luque2019impact}, which provides people a certain degree of freedom to define the specific learning objective with different statistics of the output results, like \textit{Precision} and \textit{Recall}. However, as such statistical values are always discrete, evaluation measures are insensitive to slight improvements in the model \cite{weidman2019deep}. Also, since there is a gap between observed results in existing data and the underlying posterior probability distribution \cite{kline2005revisiting}, these evaluation measures are considered theoretically incomplete in assessing the generalization ability of the model. The other type of metric is the \textbf{loss value} serving as the optimization objective in model training \cite{chen2021communication}. While researchers have proven that losses do not suffer from the issues above \cite{weidman2019deep, hung1996estimating}, they are always fixedly predefined upon the probabilistic prediction scores of all the classes, and as a result, lack the flexibility to be customized for different tasks. 

With more and more classifiers used in safety-critical areas such as self-driving \cite{soares2019actively} and healthcare \cite{lucieri2020interpretability}, just uniformly assessing different models built for different tasks is gradually found to be limited and unsatisfying. This is because, beyond the general performance, there are usually some specific \textbf{human values} to be concerned with and weighed in such areas. For instance, the mistakes to recognize a red light respectively as yellow and green have completely different degrees of severity in real-world traffic \cite{hmg2016traffic}. In this paper, we mainly focus on introducing \textbf{two} kinds of human values into multi-classifiers. Firstly, many specific cases in our human society are not just black-or-white, so instead of equally punishing all kinds of incorrect predictions with different destructiveness, we allow assigning specific weights individually for every single error case. In this way, we can specify which kinds of errors should be strictly punished, and which are relatively tolerable. Secondly, in order to satisfy some safety-critical requirements in practice, we would rather make certain compromises in overall performance to reduce specific dangerous cases. For this purpose, we introduce a threshold to indicate the concession we can make in the confidence of prediction, to leave enough space to search for model parameters that are better under the given requirements.

Based on these ideas, we propose a novel evaluation measure named \textbf{Meta Pattern Concern Score} (\textit{MPCS}) for multi-classification. We design this measure for the following targets: 1) it can be flexibly customized for different tasks to find better models given the human values mentioned; 2) it is still qualified enough to serve as a general evaluation measure, which means the models found should also be fine under assessments of the common metrics; 3) it should reduce the impact of the two inherent drawbacks of existing evaluation measures mentioned. Specifically, we take the basic idea of the confusion matrix to design an abstract representation for the probabilistic prediction result, to specify different punishments for every single kind of incorrect prediction. At the same time, through introducing the adjustable fine-grained interval into the representation, we make \textit{MPCS} approximating to \textit{cross entropy} loss in the limit. In other words, \textit{cross entropy} can be viewed as a special case of \textit{MPCS}. In this way, we not only make it inherit certain mathematical completeness and work effectively as general metrics, but also reduce the insensitivity caused by the discrete value. Finally, in addition to the evaluation, although the abstract representation makes \textit{MPCS} non-differentiable, which means it can not directly serve as the loss, we can still use it to refine the training by dynamically adjusting the learning rate according to its value.

The experiments on four different kinds of models and six reality and synthetic datasets confirm that \textit{MPCS} is as effective and computationally efficient as the common metrics in the evaluation of multi-classifiers. A specific case study in MNIST shows that given customized requirements, \textit{MPCS} can not only accordingly pick out the ideal model better than the one selected by the common metrics without violating them too much, but also improve the training process to newly train a better model. Specifically, it can pick a model with 0.53\% fewer dangerous misclassifications by just sacrificing 0.04\% of training accuracy, or directly train a new model averagely outperforming the original one with a 1.62\% lower value of \textit{MPCS} and 0.36\% fewer number of dangerous cases, which can be especially useful in real-world safety-critical applications.







\section{Background}

\subsection{Common Metrics for Classification}\label{subsec:basic}
A number of metrics have been proposed to train and pick out the desired classifier. For one thing, the evaluation measures based on simple observations and statistics of the output results such as classification accuracy rate are widely used. Most of these measures are derived from the confusion matrix, a structure to illustrate the results obtained from a classifier \cite{luque2019impact}. Given a $c$-class classification task, let $m_{ij}$ be the number of samples actually belonging to the $i$-th class while that are classified into the $j$-th class, the confusion matrix can be defined as $\mathcal{M} = \begin{bmatrix} m_{ij} \end{bmatrix}$. Given a target positive label $r$, there are four different situations namely ``True Positive'' (\textit{TP} $= m_{rr}$), ``False Negative'' (\textit{FN} $= \sum_{j=1}^{c} \hspace{0.2em} m_{rj} - m_{rr}$), ``False Positive'' (\textit{FP} $= \sum_{i=1}^{c} \hspace{0.2em} m_{ir} - m_{rr}$) and ``True Negative'' (\textit{TN} $= \sum_{i=1}^{c} \hspace{0.2em} \sum_{j=1}^{c} \hspace{0.2em} m_{ij} - \sum_{i=1}^{c} \hspace{0.2em} m_{ir} - \sum_{j=1}^{c} \hspace{0.2em} m_{rj} + m_{rr}$) \cite{ibrahim2021deep}, with which a series of evaluation measures are defined as follows:
\begin{equation*}
    \begin{aligned}
    & \textit{\textbf{Accuracy}} = \frac{\textit{TP}+\textit{TN}}{\textit{TP}+\textit{FN}+\textit{FP}+\textit{TN}}\textbf{;} \hspace{1em} \textit{\textbf{Precision}} = \frac{\textit{TP}}{\textit{TP}+\textit{FP}}\textbf{;}\\
    & \textit{\textbf{Recall}} = \frac{\textit{TP}}{\textit{TP}+\textit{FN}}\textbf{;} \hspace{1em} \textbf{F}_{1} \textbf{\text{-}} \textit{\textbf{score}} = 2 \times \frac{\textit{Precision} \times \textit{Recall}}{\textit{Precision}+\textit{Recall}} \hspace{0.1em} \textbf{;}\\
    & \textit{\textbf{MCC}} = \frac{\textit{TP} \times \textit{TN} - \textit{FP} \times \textit{FN}}{\sqrt{(\textit{TP}+\textit{FP})(\textit{TP}+\textit{FN})(\textit{TN}+\textit{FP})(\textit{TN}+\textit{FN})}} \hspace{0.1em} \textbf{.}\\
    \end{aligned}
\end{equation*}

For another, the objective of model training is to find the optimal parameters $\omega^{*}$ to minimize the value of global loss function $\mathcal{L}(\omega)$ over the whole training dataset $\mathcal{X}$ \cite{chen2021communication}:
\begin{equation}
    \begin{aligned}
    \omega^{*} = arg \hspace{0.2em} min \hspace{0.2em} \mathcal{L}(\omega) = arg \hspace{0.2em} min \hspace{0.2em} \frac{1}{|\mathcal{X}|} \sum_{x \in \mathcal{X}} l(x,\omega)
    \end{aligned}
\end{equation}
where $l(x,\omega)$ denotes the specific loss value computed from sample $x$ with parameters $\omega$. So naturally, how well the objective is met can serve as a metric of model quality. Given a model $\mathcal{N}$ having fully connected final layer with the \textit{softmax} activation, the supervised labels $y$ in the format of one-hot, and the probability outputs $\hat{y}$ predicted by $\mathcal{N}$, the famous loss \textit{mean squared error} (\textit{MS}) and the \textit{cross entropy error} (\textit{CE}) commonly used in classification \cite{das2019separability} can be denoted as:
\begin{equation}
    \begin{aligned}
    \label{Equ:mean squared error}
    \mathcal{L}_{\textit{MS}} = -\frac{1}{2|\mathcal{X}|} \sum_{i=1}^{|\mathcal{X}|} \hspace{0.2em} \sum_{j=1}^{c} \hspace{0.2em} (\hat{y}_{i}^{(j)} - y_{i}^{(j)})^2
    \end{aligned}
\end{equation}
\begin{equation}
    \begin{aligned}
    \label{Equ:cross entropy error}
    \mathcal{L}_{\textit{CE}} = -\frac{1}{|\mathcal{X}|} \sum_{i=1}^{|\mathcal{X}|} \hspace{0.2em} \sum_{j=1}^{c} \hspace{0.2em} y_{i}^{(j)} \hspace{0.2em}  log(\hat{y}_{i}^{(j)})
    \end{aligned}
\end{equation}

\subsection{Drawbacks of Existing Metrics}
The two kinds of metrics mentioned above both have their own advantages in contrast, while they also struggle with different dilemmas. Compared with loss value, confusion matrix-based evaluation measures have advantages in the ability of flexible customization. Through the various combinations of \textit{TP}, \textit{FN}, \textit{FP} and \textit{TN}, evaluation measures allow one to define the objective in an intuitive and direct format according to the specific application. For instance, as the over-abundance of negative examples is commonly seen in information retrieval, recommendation systems and social network analysis \cite{flach2015precision}, one of the reasonable metrics in such situations is the \textit{Precision} who can measure the ability of the classifier not to label as positive a sample that is negative, so that the always-negative classifier will not be over-valued.

Nevertheless, evaluation measures directly refer to the classification results, which brings some potential problems. For one thing, each element in the confusion matrix is essentially a number of classification results having specific actual and predicted labels, which means the values of all the evaluation measures based on the confusion matrix are discrete. As a consequence, they are insensitive to slight improvements in model parameters and only vary discontinuously and abruptly when the prediction results of some samples change \cite{weidman2019deep}. This can be especially disappointing in the latter stages of model training. For another, evaluation measures heavily focus on the existing data, while the ideal model is expected to reliably classify previously unseen objects in the real world. So what really matters is, according to the \textit{Bayesian} classification theory, how well the model approximates the underlying posterior probability distribution \cite{duda1973pattern}. Researchers have shown that improving observed accuracy rates and improving posterior probability estimation are not entirely synonymous \cite{kline2005revisiting}. In fact, efforts to improve posterior probability estimation may yield lower accuracy rates in the known dataset, but lead to better performance on future reality tasks.

The problems mentioned above are actually the important reasons why loss is proposed \cite{weidman2019deep}. Carefully designed loss functions ensure the gradient of most parameters is not $\vec{\textbf{0}}$ in the training process, so as to guide the continuous optimization of the model. And for the latter problem mentioned, the theoretical relationship between \textit{Bayesian} posterior probability estimation and \textit{MS} cost functions had also been explored and widely adopted \cite{hung1996estimating}. However, squared error assumes \textit{Gaussian} target data, which is violated given the discrete targets used to train classifiers. Besides, it can be found from \eqref{Equ:mean squared error} that \textit{MS} is likely to be dominated by a few outlier data points that have particularly large errors in practice.

As a \textit{log}-linear error function based on the maximum likelihood estimate approach \cite{das2019separability} that is impacted much less by these further problems, \textit{CE} gradually became more popular in classification tasks to date. However, entropy-based measures are specifically designed for binary targets at the beginning, so when \textit{CE} is applied in multi-classification tasks, it has to treat all the classes except the class with the correct label as a collective concept ``incorrect class'', without considering the distinction between them. What's worse, \textit{CE} merely takes the probability of the correct label in the prediction results into error calculation, which means that of the incorrect labels, which are certainly more in quantity in every single output of multi-classification and may also impact the prediction effectiveness, are totally ignored.

All in all, there is indeed an uncharted territory: is the bias between the two kinds of metrics irreconcilable? Empirically, if we consider the multi-classification task from a similar perspective as the confusion matrix, but construct a specific formula with the mathematical principle closer to the loss, it is likely that a metric can be proposed with not only the customizability just as the former but also the theoretical completeness inherited from the latter at the same time. We take this idea into consideration when specifically designing the formula of our own evaluation measure.

\section{Meta Pattern Concern Score}\label{sec:mainApp}

In the past section, from the perspective of basic technology, we discussed the ideas and issues to be concerned with for our design. In this section, we first rely on two real-world scenarios to illustrate and describe the motivation and intuitive scheme to take the two kinds of human values into consideration, and then accordingly propose our novel evaluation measure, \textbf{Meta Pattern Concern Score} (\textit{MPCS}).

\subsection{Considering Human Values in \textit{MPCS}}
The two scenarios to be talked about come from a typical safety-critical task, Traffic Light Recognition, which is important regarding the traffic participants' safety in autonomous driving \cite{fernandez2018deep}. Note that this task usually involves a variety of technologies in practice, but for convenience here we just simplify it as a 3-class classification problem.

The first scenario happens when a red light shows, where there are two different kinds of misclassifications according to the specific prediction result. If it is classified as a green light, the autonomous vehicle will be incorrectly allowed to proceed, which may endanger both its passengers and others, and as a result, totally unacceptable. While if the light is predicted to be yellow, which means ``stop, unless it is unsafe to do so'' \cite{hmg2016traffic}, at least in most cases the car will stop and not break the traffic rules. So it is obviously inappropriate to treat both of the errors with the same severity, which is, however, what the traditional metrics do.

\begin{figure}[htb]
    \centering
    \includegraphics[width=5.5cm]{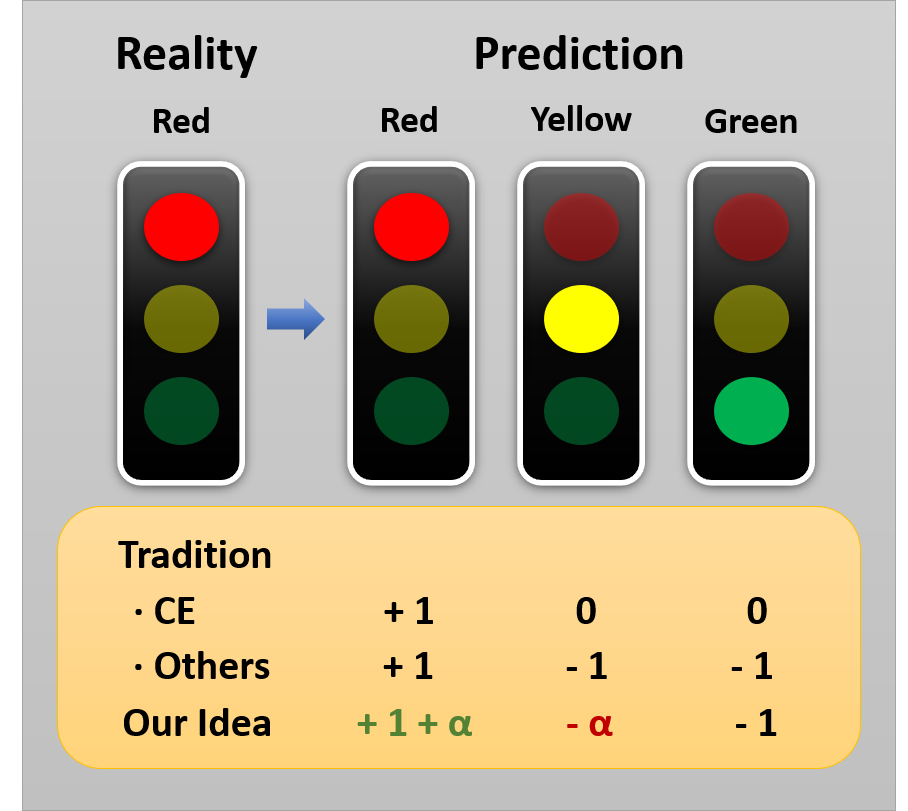}
    \caption{The figure shows the simplified traffic light classification problem. The values in the figure illustrate how the three different predictions (i.e. predicting ``Red'' respectively as ``Red'', ``Yellow'' and ``Green'') contribute to the calculation of different metrics. This corresponds to a concept to be formally defined later named \textbf{concern degree}. The ``Others'' includes the \textit{MS} and all the evaluation measures from the confusion matrix.}
    \label{fig:example1}
\end{figure}

Since this kind of scenario with negative classes having different damages is commonly seen in practice, it would make sense to design for it. But before providing formal definitions and descriptions in section \ref{sec:Approach}, showing our idea in an intuitive way firstly might be found helpful. As mentioned previously, \textit{CE} is only affected by the predicted value of the correct label. Just as shown in Fig.~\ref{fig:example1}, this is marked as follows: the significance of the correct class ``Red'' is $+1$, while that of others is assigned to be $0$ to represent no influence. On the other hand, the \textit{MS} and all the evaluation measures from confusion matrix treat the prediction as either True or False, respectively corresponding to $+1$ and $-1$ in the ``Others'' line. 

In our idea, we mark $-\alpha$ instead of $-1$ for the incorrect prediction ``Yellow'' to represent it is relatively less destructive. What's more, we assign that of the ``Red'' to be $-((-\alpha)+(-1))=1+\alpha$, which is basically a compromise between \textit{CE} and other metrics. Specifically, we neither ignore negative classes completely nor make each of them as important as the positive class, while we set that the positive class and the whole of negative classes are of equal significance. These marked values correspond to a concept to be formally defined later in our measure called \textbf{concern degree}.

\begin{table*}[htbp]
\caption{A simple case showing that the classifier with a lower loss value is not necessarily the better one in specific safety-critical tasks. Specifically, classifier \textit{A} correctly predicts 99 samples with the confidence of 0.99 and wrongly predicts one sample with a confidence of 0.49, while classifier B correctly predicts all the 100 samples with the confidence of 0.98.}
\label{tab:example2}
\begin{center}
\renewcommand{\arraystretch}{1.2}
\resizebox{15cm}{!}{
\begin{tabular}{|c|c|c|c|c|c|}
\hline
\textbf{Training Size} & \textbf{Classifier} & \textbf{Prediction Result} & \textbf{Number of Sample} & \textbf{Prediction Confidence} & \textbf{Cross Entropy Loss} \\
\hline
\multirow{4}{*}{100} & \multirow{2}{*}{\textbf{A}} & \color{teal}{Correct} & \color{teal}{99} & \color{teal}{0.99} & \multirow{2}{*}{0.741920} \\
\cline{3-5} 
& & \color{purple}{Wrong} & \color{purple}{1} & \color{purple}{0.49} & \\
\cline{2-6} 
& \multirow{2}{*}{\textbf{B}} & \color{teal}{Correct} & \color{teal}{100} & \color{teal}{0.98} & \multirow{2}{*}{0.877392} \\
\cline{3-5} 
& & \color{purple}{Wrong} & \color{purple}{0} & \color{purple}{-} & \\
\hline
\end{tabular}
}
\end{center}
\end{table*}

\begin{figure*}[htb]
    \centering
    \includegraphics[width=11cm]{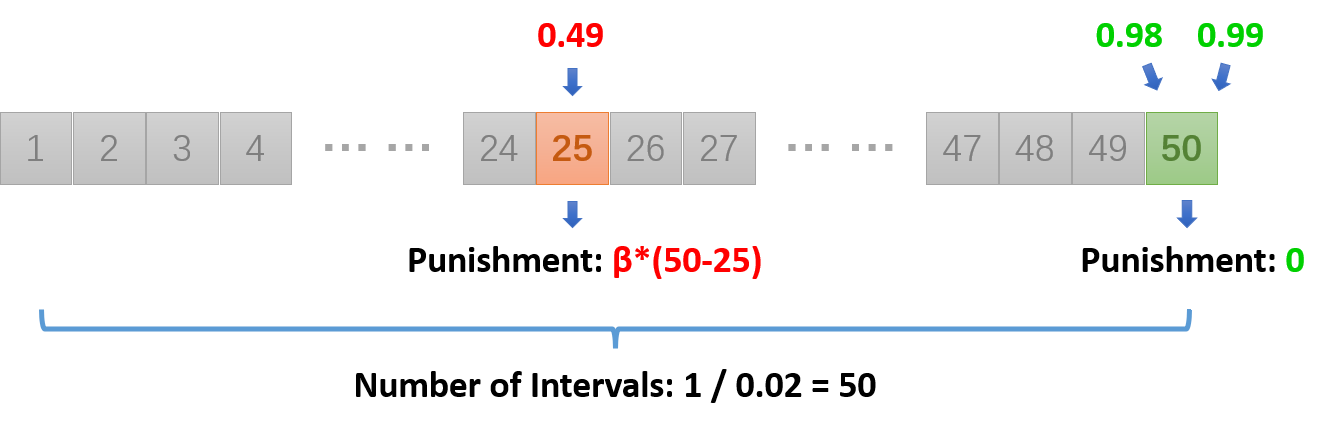}
    \caption{The figure illustrates our idea to set a threshold for prediction confidence (e.g. 1 - 0.02 = 0.98) to indicate the concession we can tolerate, and divide the confidence field into different intervals to extend the concession interval by interval, to accordingly calculate the specific punishment values.}
    \label{fig:example3}
\end{figure*}

The second scenario is presented aiming to illustrate the significance to introduce a threshold for the concession in prediction confidence. As shown in Table~\ref{tab:example2}, there are two classifiers trained for a task with a training size of 100 and the loss \textit{CE}. The classifier \textit{A} makes a mistake in one of the input samples, predicting the probability of the correct label to be only $0.49$, while the classifier B makes correct predictions in the whole dataset. However, the confidence of all the correct predictions of classifier \textit{A} is $0.99$, while that of \textit{B} is $0.98$. In this case, it is indeed that the \textit{CE} value of classifier \textit{A} is less than that of \textit{B}, but does that really mean \textit{A} is a better choice? At least in our simplified Traffic Light Recognition case, the answer is no because it is the failure of prediction that is more intolerable compared with the minor weakness of confidence in such a safety-critical task. Note that although here we take \textit{CE} as an instance, \textit{MS} and other loss functions also have the same problem to different extents.

To deal with this problem, we can set a threshold value for the confidence and keep the model from being punished in specific samples once the confidence value of the correct label exceeds it. In our idea shown in Fig.~\ref{fig:example3}, we further divide the entire confidence field into several intervals according to the granularity determined by the threshold, and calculate the punishment of a specific confidence value based on the distance between the interval it falls into and the interval representing the highest confidence.

\subsection{Detailed Design of \textit{MPCS}}\label{sec:Approach}
\begin{figure*}[htb]
    \centering
    \includegraphics[width=18cm]{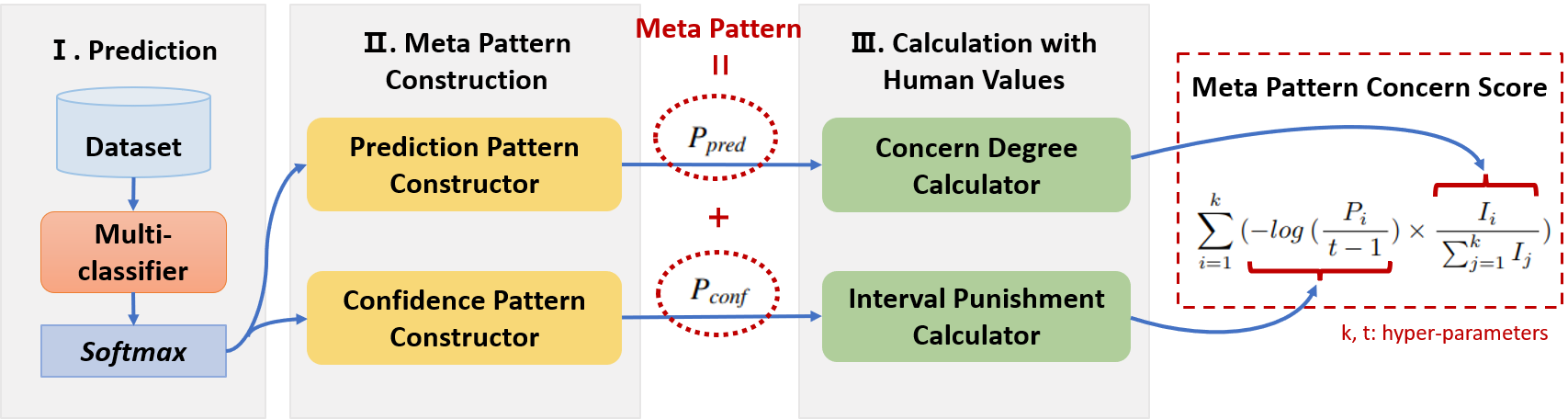}
    \caption{The figure shows the calculation process of \textit{MPCS}. The meta pattern includes the prediction pattern and the confidence pattern constructed from probabilistic predictions of the target multi-classifier. Given specific human values, the concern degrees and interval punishments are calculated upon meta pattern, and they are then used to calculate the final \textit{MPCS} value.}
    \label{fig:roadmap}
\end{figure*}

In this section, we propose the specific definition of \textit{MPCS}, and provide a detailed process to calculate its value. As shown in Fig.~\ref{fig:roadmap}, the calculation process can be divided into three parts. The first part is about running the multi-classifier and normalizing its results to the probabilistic distribution using \textit{softmax}. The second part is where the two constructors extract and build meta patterns from the processed results. Finally, in the third part, concern degrees and interval punishments are calculated given specific human values, and the \textit{MPCS} on the entire dataset is accordingly acquired.

To begin with, to realize our idea, it is not sufficient enough to merely take the label with the highest confidence into account. However, considering all the labels may also lead to a lot of unnecessary computational time costs. So we introduce a hyper-parameter $k$ to indicate the top-$k$ classes in the order of confidence decreasing to be considered according to the practical task. With $l_{i}$ denoting a specific label in a classification task, the \textbf{prediction pattern} is defined as a vector $\textit{P}_{\textit{pred}}$ with length $k$: 
\begin{equation}
    \begin{aligned}
    \textit{P}_{\textit{pred}}=[l_{1},l_{2},\dots,l_{k}]
    \end{aligned}
\end{equation}
where the value of $k$ is between one and the number of classes of the task.

To realize the idea described in Fig.~\ref{fig:example3}, we divide the confidence field into $t$ intervals and determine the confidence level of a specific label under an input sample by the interval in which its confidence value falls and whether the label is the correct one or not. The \textbf{confidence pattern} is defined as a vector $\textit{P}_{\textit{conf}}$ with length $k$ that must be the same as $\textit{P}_{\textit{pred}}$ in one task:
\begin{equation}
    \begin{aligned}
    \textit{P}_{\textit{conf}}=[C(l_1),C(l_2),\dots,C(l_k)]
    \end{aligned}
\end{equation}
in which $C(l_i)$ denotes the confidence level of the label $l_i$. Given a hyper-parameter $t \geq 1$ and the correct label $l$, and let $c_i$ be the confidence of label $l_i$, $C(l_i)$ can be calculated as follows:
\begin{equation}
    \begin{aligned}
    C(l_i)=
    \begin{cases}
        \lfloor t \cdot c_i \rfloor, & l_i = l\\
        \hspace{0.2em} t - \lfloor t \cdot c_i \rfloor - 1, & l_i \neq l\\
    \end{cases}
    \end{aligned}
\end{equation}

Notice that there is a correspondence between each pair of elements in the same position of $\textit{P}_{\textit{pred}}$ and $\textit{P}_{\textit{conf}}$ in any specific sample. As a pair of the two patterns can be used to record the key information to uniquely represent any specific output of the classifier, we always refer to them together. The \textbf{meta pattern} is an abstract representation of the probabilistic prediction result of multi-classifiers, which is defined as the combination of the corresponding prediction pattern $\textit{P}_{\textit{pred}}$ and confidence pattern $\textit{P}_{\textit{conf}}$.

Then it is time to recall the idea shown in Fig.~\ref{fig:example1}. We introduce a release list as a hyper-parameter with every single element indicating a less destructive situation in practice, as well as a factor to indicate that in what extent the concern of them could be released. In every single data sample, for every label selected into its $\textit{P}_{\textit{pred}}$, there are three kinds of values of the corresponding element in its concern degree depending on whether the label is the correct one and whether it exists in any element of release list together with the correct label if not. Given a prediction pattern $\textit{P}_{\textit{pred}}$, its \textbf{concern degree} can be defined as a vector $\mathcal{I}$ with the same length $k$:
\begin{equation}
    \begin{aligned}
    \mathcal{I}=[D(l_1),D(l_2),\dots,D(l_k)]
    \end{aligned}
\end{equation}
Given the correct label $l$, a release list $\mathcal{R}$ with $r_{\alpha}$ representing any element of it, and a release factor $f_{\mathcal{R}}$, $D(l_i)$ can be calculated as:
\begin{equation}
    \begin{aligned}
    D(l_i)=
    \begin{cases}
        \sum_{l_j \in \textit{P}_{\textit{pred}}, \hspace{0.2em} j \neq i} D(l_j), & l_i = l\\
        \hspace{0.2em} f_{\mathcal{R}}, & l_i \neq l, \hspace{0.2em} [l,li] \in r_{\alpha}\\
        \hspace{0.2em} 1, & l_i \neq l, \hspace{0.2em} [l,li] \notin r_{\alpha}\\
    \end{cases}
    \end{aligned}
\end{equation}

\begin{algorithm}[t]
\caption{Calculation of Meta Pattern Concern Score}
\label{alg:algorithm}
\textbf{Input}: Classifier $\mathcal{N}$, Dataset $(\mathcal{X},\mathcal{Y})$, Release List $\mathcal{R}$, Release Factor $f_{\mathcal{R}}$, Hyper-parameters $k$ and $t$\\
\textbf{Output}: Meta Pattern Concern Score $\mathcal{S}$

\begin{algorithmic}[1] 
\STATE $\mathcal{S}' \leftarrow 0$
\WHILE{$x,y \in (\mathcal{X},\mathcal{Y})$}
\STATE $P_{\textit{pred}} \leftarrow argsort(-\mathcal{N}(x))[:k]$
\STATE $P_{\textit{conf}} \leftarrow t - floor((-sort(-\mathcal{N}(x))[:k]) \times t) - 1$
\STATE $P_{\textit{conf}}[P_{\textit{conf}}=-1] \leftarrow 0$
\STATE $\mathcal{I} \leftarrow ones(k)$
\IF {\textbf{exist} $i \leftarrow P_{\textit{pred}}.index(y)$}
\STATE $P_{\textit{conf}}[i] \leftarrow t - P_{\textit{conf}}[i] - 1$
\WHILE{$r \in \mathcal{R}$ \textbf{and} $P_{\textit{pred}}[i] = r[0]$}
\WHILE{$j$ \textbf{from} $0$ \textbf{to} $k-1$}
\IF {$j \neq i$ \textbf{and} $P_{\textit{pred}}[j] \in r[1:]$}
\STATE $\mathcal{I}[j] \leftarrow f_{\mathcal{R}}$
\ENDIF
\ENDWHILE
\ENDWHILE
\STATE $\mathcal{I}[i] \leftarrow sum(\mathcal{I}) - 1$
\ENDIF
\STATE $P_{\textit{conf}}[P_{\textit{conf}}=0] \leftarrow \textbf{1e-7}$
\STATE $P' \leftarrow -log(P_{\textit{conf}}/(t-1))$
\STATE $\mathcal{I}' \leftarrow \mathcal{I} / sum(\mathcal{I})$
\STATE $\mathcal{S}' \leftarrow \mathcal{S}' + sum(multiply(P', \mathcal{I}'))$
\ENDWHILE
\STATE $\mathcal{S} \leftarrow \mathcal{S}' / |\mathcal{X}|$
\STATE \textbf{return} $\mathcal{S}$
\end{algorithmic}
\end{algorithm}

\begin{table*}[htbp]
\caption{The Spearman similarity between \textit{MPCS} and the five benchmark metrics \textit{Accuracy} (\textit{ACC}), $F_{1} \text{-}$\textit{score} (\textit{F1}), \textit{MCC}, \textit{MS} and \textit{CE} under the four models and six datasets. The high similarities confirm that \textit{MPCS} is effective as benchmark metrics in general tasks.}
\label{tab:Similarity}
\begin{center}
\renewcommand{\arraystretch}{1.2}
\resizebox{13cm}{!}{
\begin{tabular}{|c|c|c|c|c|c|c|c|}
\hline
\multirow{2}{*}{\textbf{ID}} & \multirow{2}{*}{\textbf{Model}} & \multirow{2}{*}{\textbf{Dataset}} & \multicolumn{5}{|c|}{\textbf{Spearman Similarity}} \\
\cline{4-8} 
& & & \textbf{ACC} & \textbf{F1} & \textbf{MCC} & \textbf{MS} & \textbf{CE} \\
\hline
\bfseries Exp. 1 & \multirow{3}{*}{MLP} & IRIS & -0.9091 & -0.9092 & -0.9011 & \bfseries 0.9873 & \bfseries 0.9827\\
\cline{1-1} \cline{3-8} 
\bfseries Exp. 2 & & DIGITS & \bfseries -0.9886 & \bfseries -0.9936 & \bfseries -0.9935 & \bfseries 0.9989 & \bfseries 0.9989\\
\cline{1-1} \cline{3-8} 
\bfseries Exp. 3 & & MNIST & \bfseries -0.9881 & \bfseries -0.9886 & \bfseries -0.9880 & \bfseries 0.9880 & \bfseries 0.9841\\
\hline
\bfseries Exp. 4 & \multirow{2}{*}{CNN} & DIGITS & -0.9106 & -0.9106 & -0.9107 & 0.9547 & 0.9547\\
\cline{1-1} \cline{3-8} 
\bfseries Exp. 5 & & MNIST & -0.9231 & -0.9229 & -0.9220 & 0.9299 & 0.9168\\
\hline
\bfseries Exp. 6 & \multirow{2}{*}{RNN} & UCR-CT & -0.9576 & -0.9549 & -0.9587 & 0.9618 & 0.9619\\
\cline{1-1} \cline{3-8} 
\bfseries Exp. 7 & & UCR-SS & \bfseries -0.9846 & \bfseries -0.9844 & \bfseries -0.9827 & \bfseries 0.9929 & \bfseries 0.9928\\
\hline
\bfseries Exp. 8 & \multirow{3}{*}{\hspace{0.5em} LSTM \hspace{0.5em}} & \hspace{0.5em} UCR-SS \hspace{0.5em} & \hspace{0.5em} -0.9382 \hspace{0.5em} & \hspace{0.5em} -0.9352 \hspace{0.5em} & \hspace{0.5em} -0.9373 \hspace{0.5em} & \hspace{0.5em} \bfseries 0.9700 \hspace{0.5em} & \hspace{0.5em} \bfseries 0.9718 \hspace{0.5em} \\
\cline{1-1} \cline{3-8} 
\bfseries Exp. 9 & & ADD & \bfseries -0.9721 & \bfseries -0.9735 & \bfseries -0.9706 & \bfseries 0.9815 & \bfseries 0.9862\\
\cline{1-1} \cline{3-8} 
\hspace{0.5em} \bfseries Exp. 10 \hspace{0.5em} & & MNIST & \bfseries -0.9757 & \bfseries -0.9754 & \bfseries -0.9760 & \bfseries 0.9774 & \bfseries 0.9763\\
\hline
\end{tabular}
}
\end{center}
\end{table*}

\begin{figure*}[htb]
\subfigure
{
    \begin{minipage}[b]{.16\linewidth}
        \includegraphics[scale=0.25]{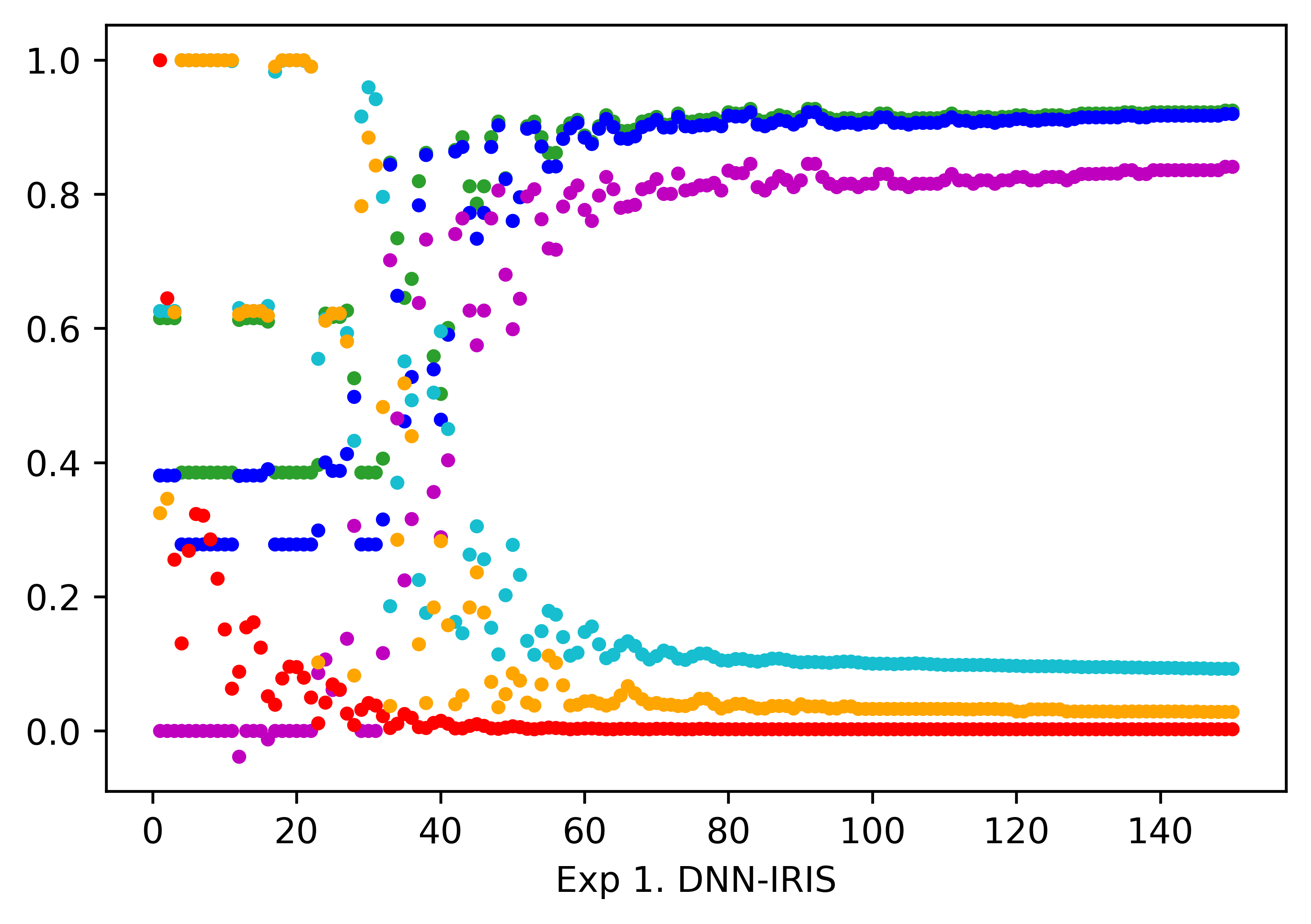}
    \end{minipage}
}
{
    \begin{minipage}[b]{.16\linewidth}
        \includegraphics[scale=0.25]{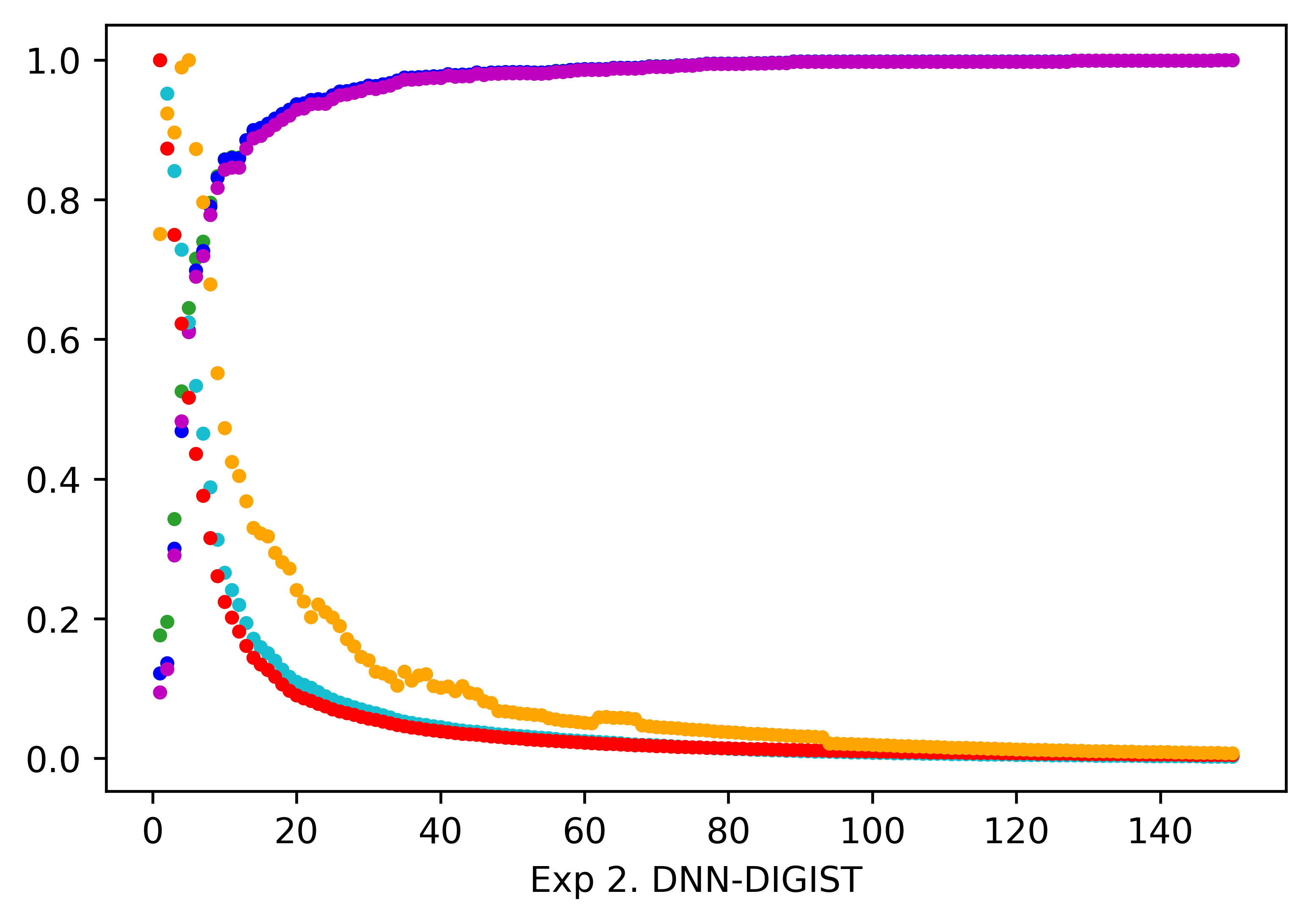}
    \end{minipage}
}
{
    \begin{minipage}[b]{.16\linewidth}
        \includegraphics[scale=0.25]{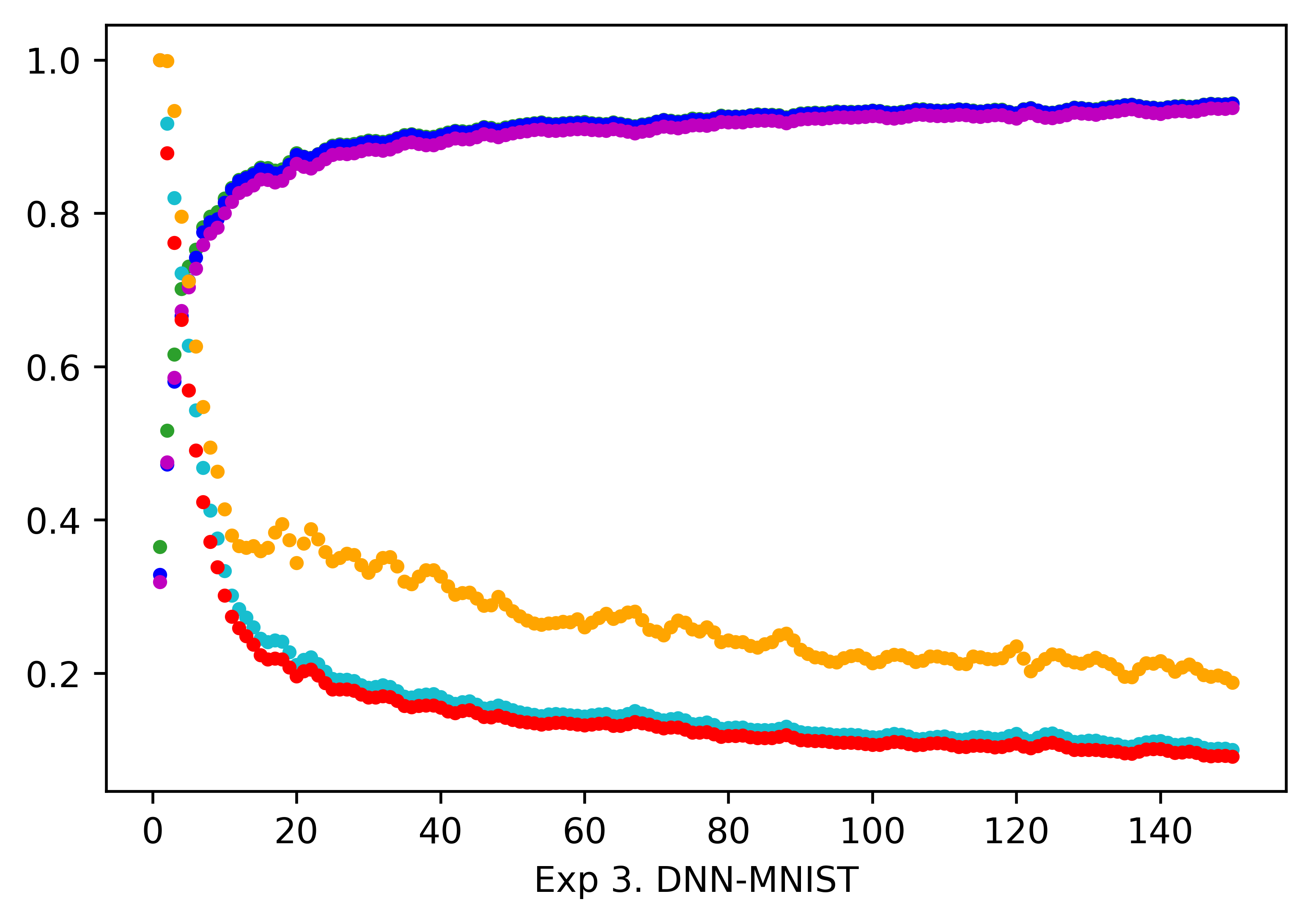}
    \end{minipage}
}
{
    \begin{minipage}[b]{.16\linewidth}
        \includegraphics[scale=0.25]{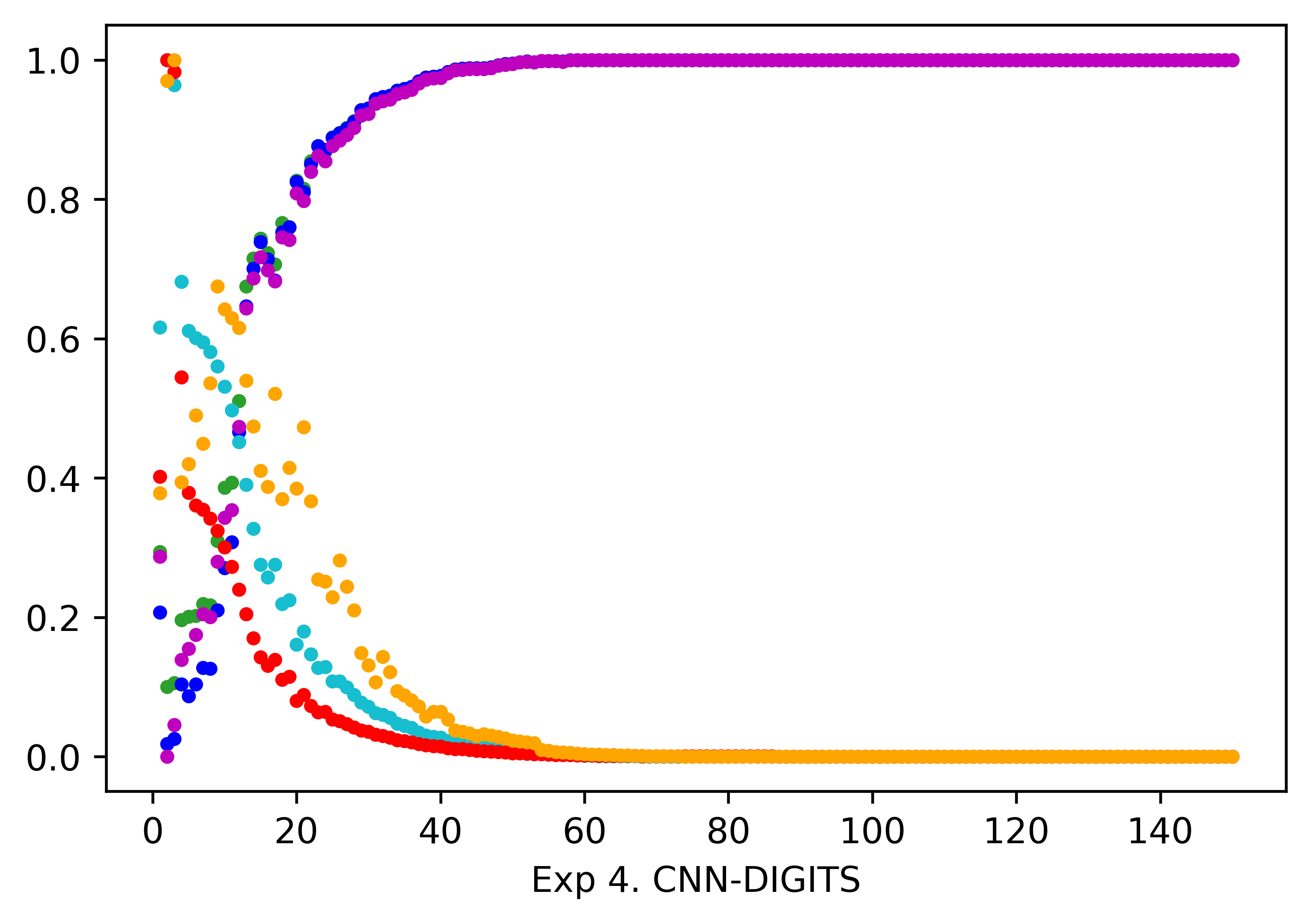}
    \end{minipage}
}
{
    \begin{minipage}[b]{.23\linewidth}
        \includegraphics[scale=0.25]{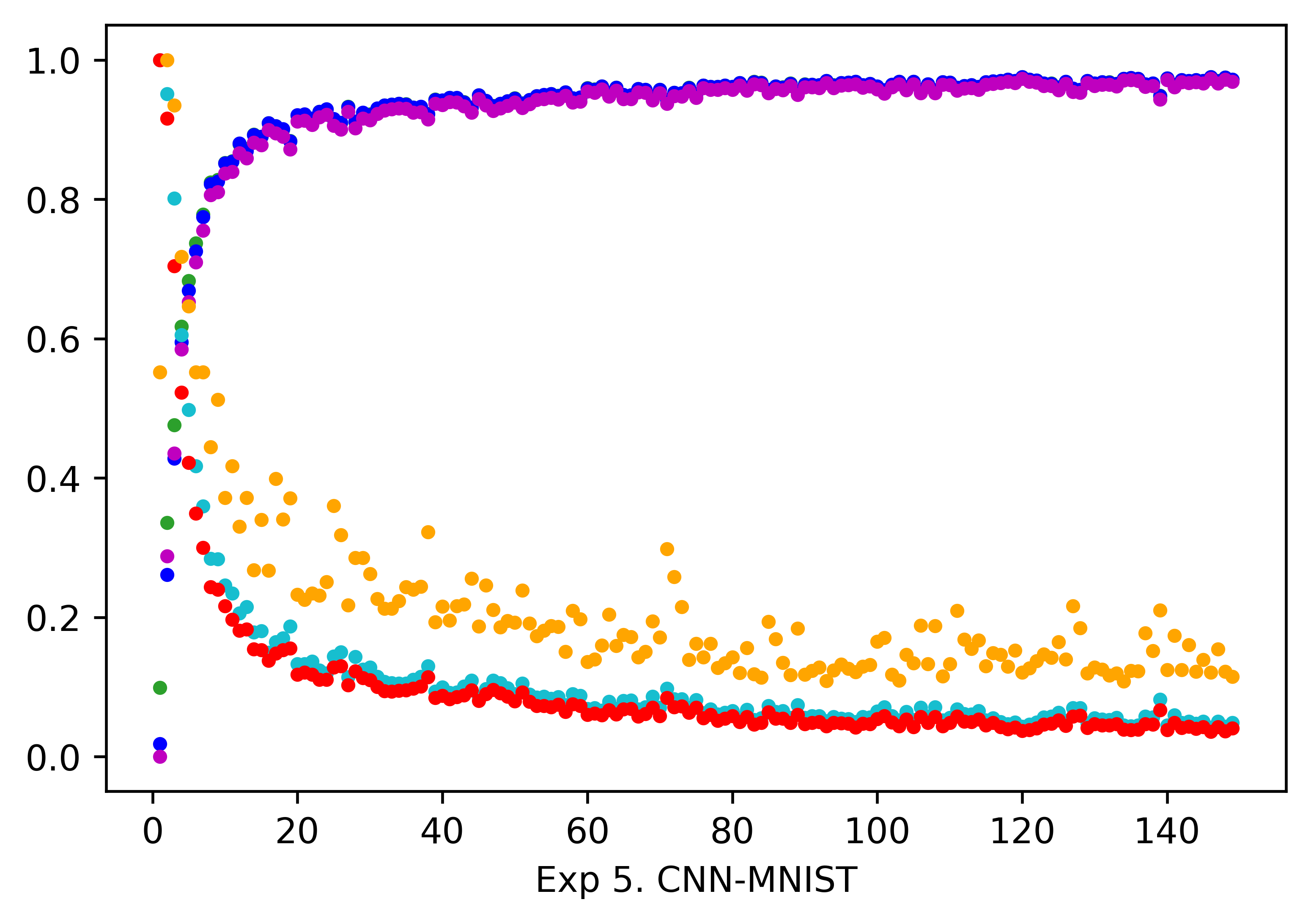}
    \end{minipage}
}
\subfigure
{
    \begin{minipage}[b]{.17\linewidth}
        \includegraphics[scale=0.25]{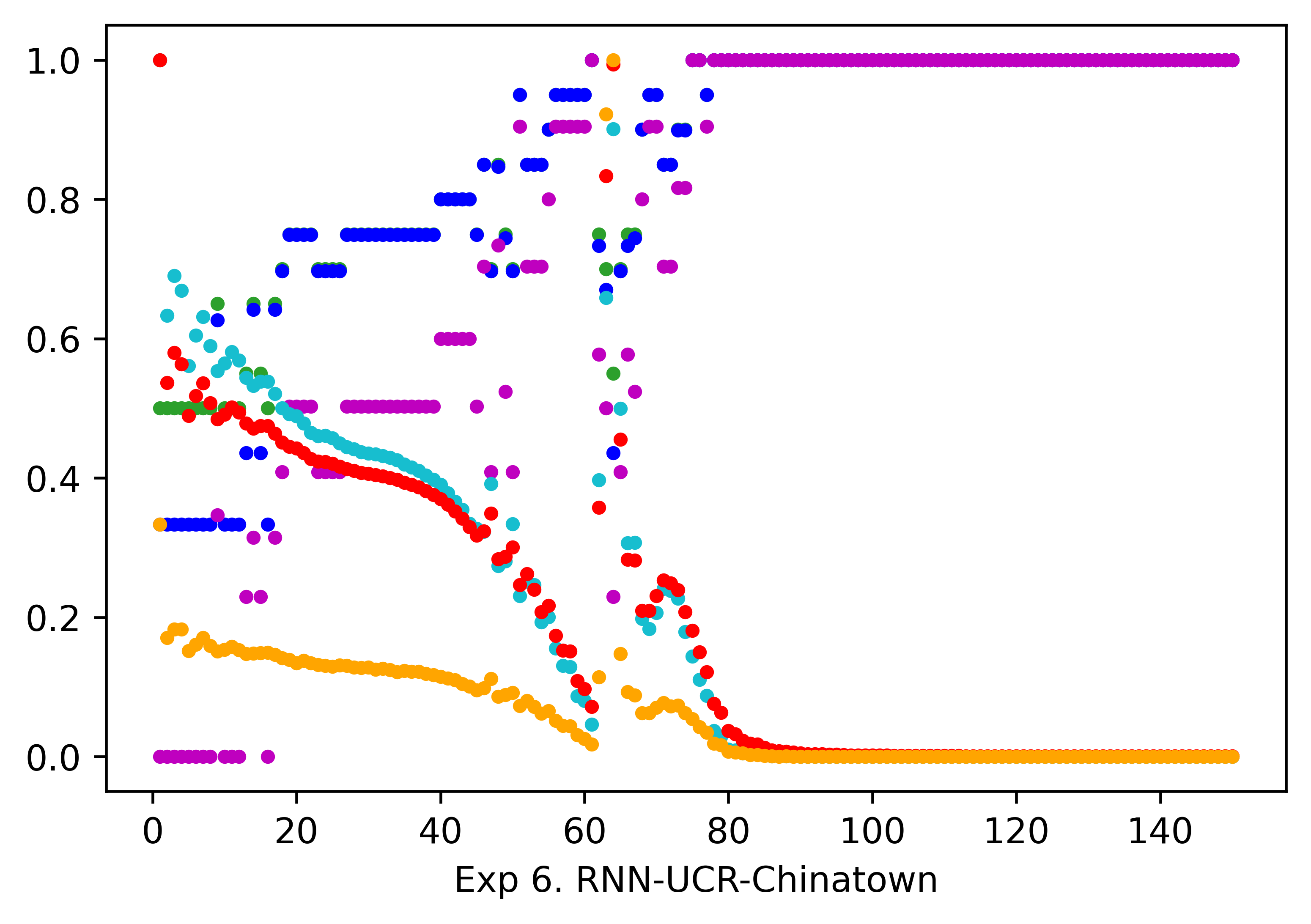}
    \end{minipage}
}
{
    \begin{minipage}[b]{.17\linewidth}
        \includegraphics[scale=0.25]{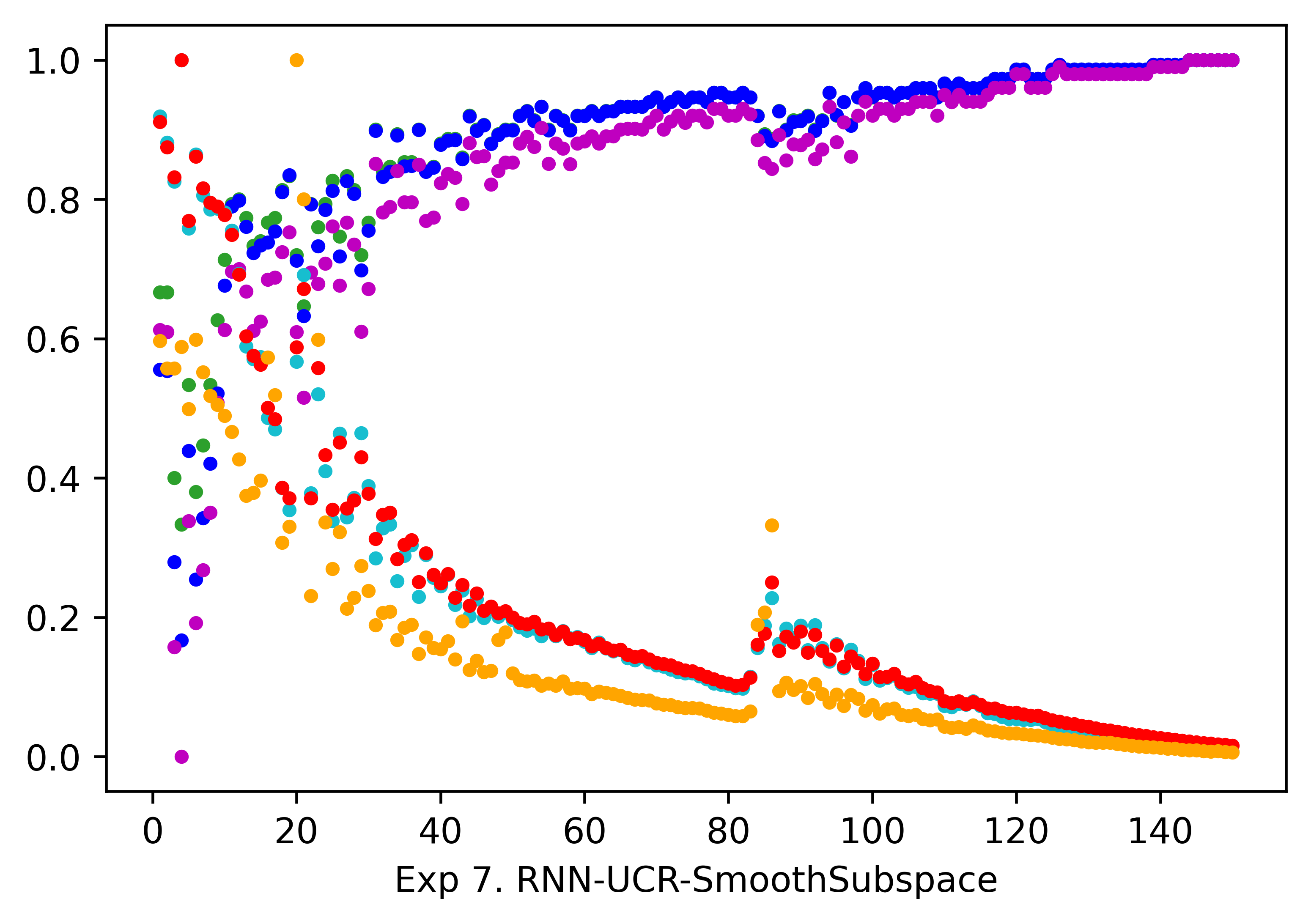}
    \end{minipage}
}
{
    \begin{minipage}[b]{.17\linewidth}
        \includegraphics[scale=0.25]{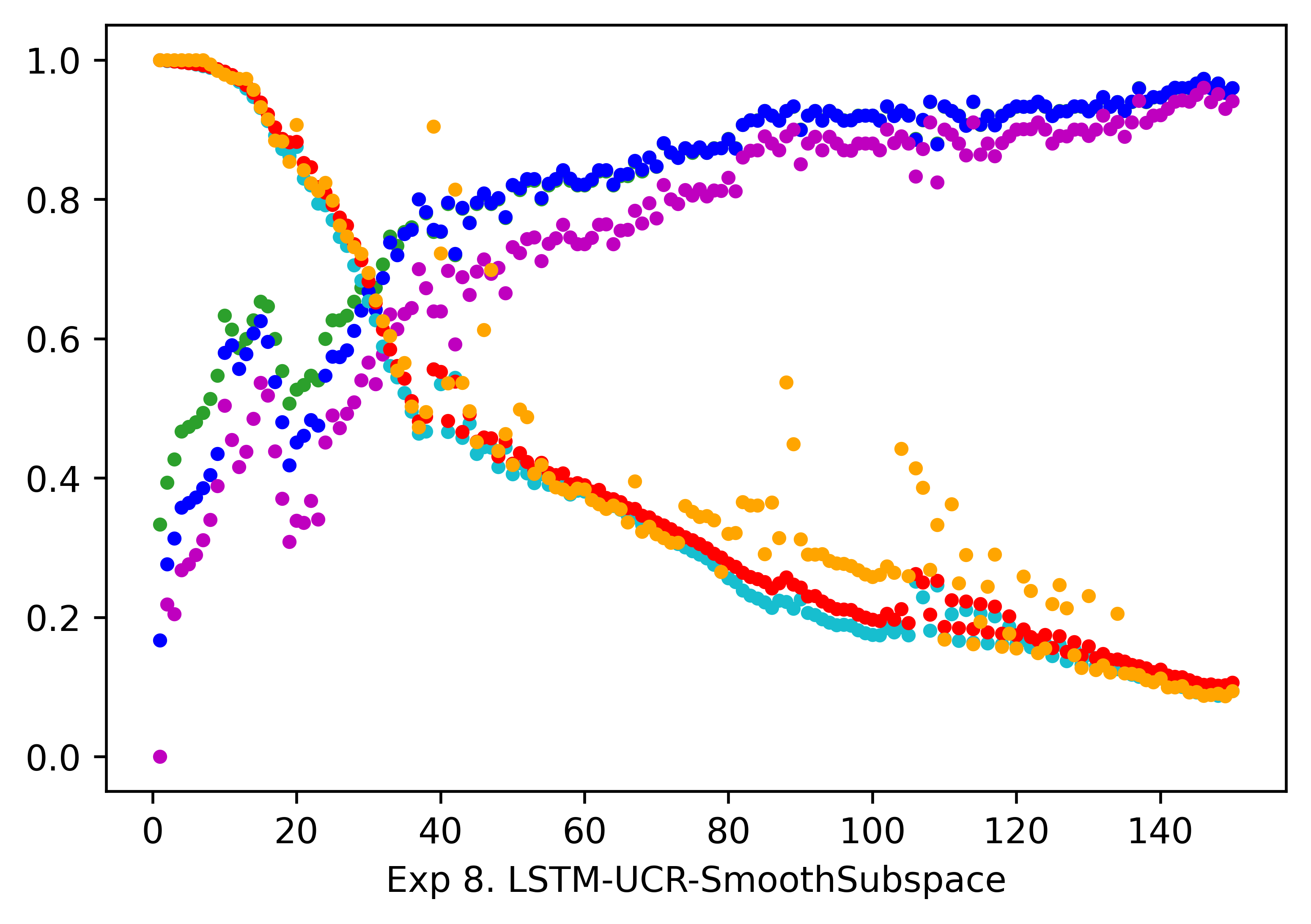}
    \end{minipage}
}
{
    \begin{minipage}[b]{.17\linewidth}
        \includegraphics[scale=0.25]{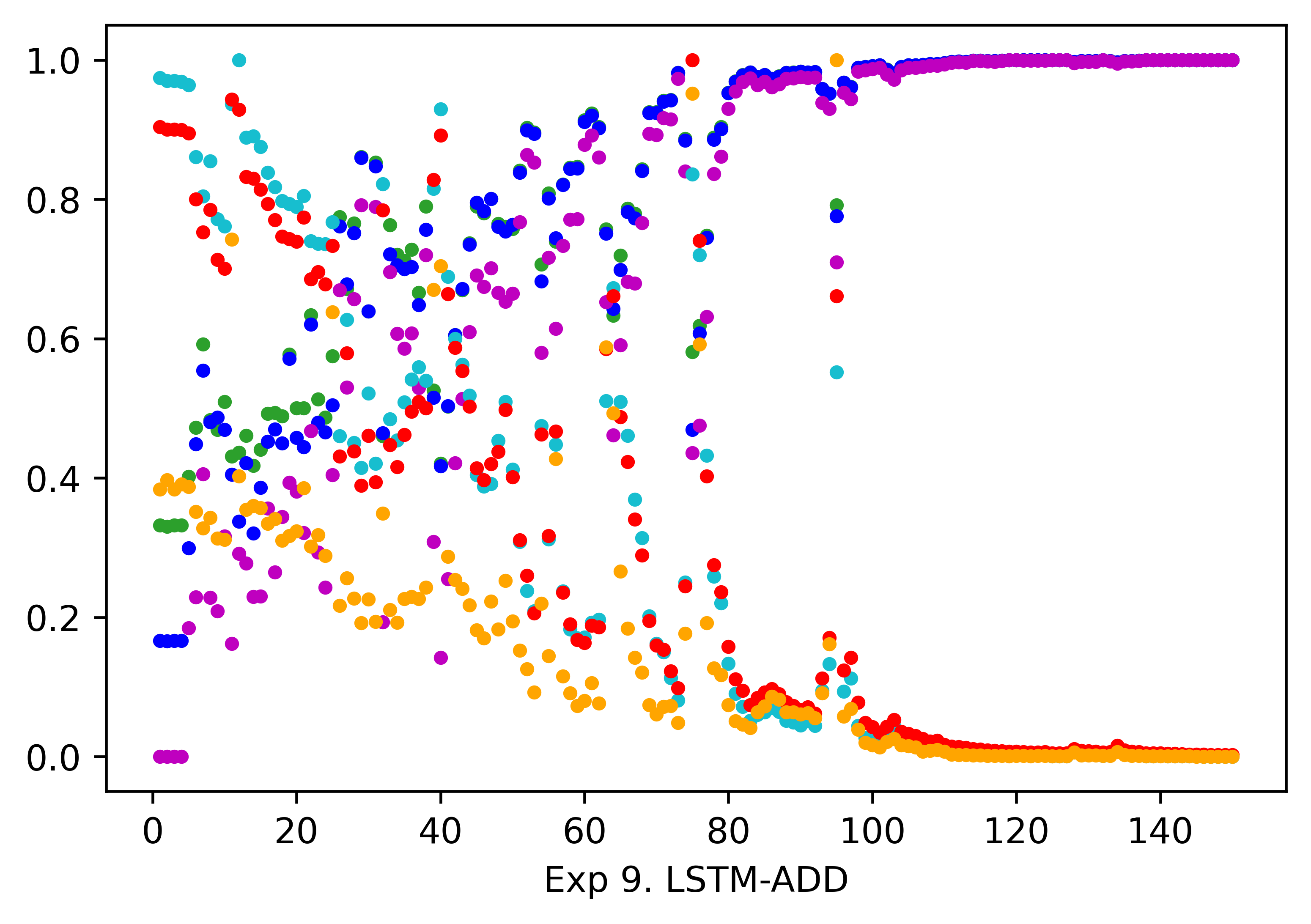}
    \end{minipage}
}
{
    \begin{minipage}[b]{.17\linewidth}
        \includegraphics[scale=0.25]{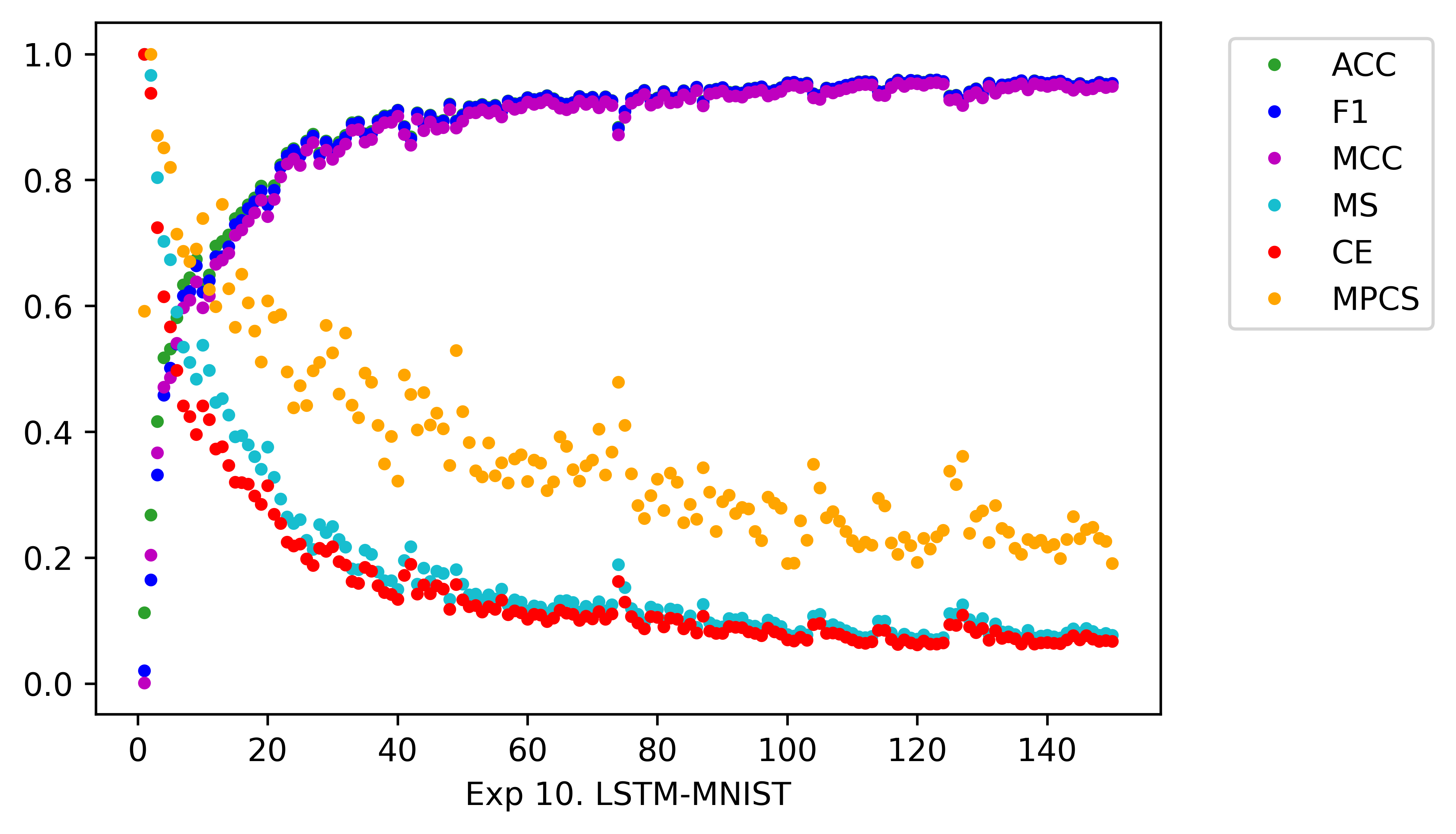}
    \end{minipage}
}
\caption{Corresponding to Table~\ref{tab:Similarity}, the figures show the trend of \textit{MPCS} and the benchmark metrics in the ten different training processes, which also intuitively show their similarity in general practice. Notice that for better visual comparison between them, the specific values of \textit{MS}, \textit{CE} and \textit{MPCS} are normalized.}
\label{fig:10}
\end{figure*}

Now we can propose our evaluation measure \textit{MPCS} based on the above definitions. To calculate the \textbf{interval punishment}, given confidence pattern $\textit{P}_{\textit{conf}}$, we select the $log$ function just as \textit{CE} to convert the interval distance into a concrete punishment value. Then with the concern degree $\mathcal{I}$, the \textbf{meta pattern concern score} $\mathcal{S}$ of the classification result of every single sample can be calculated as:
\begin{equation}
    \begin{aligned}
    \mathcal{S} = \sum_{i=1}^{k} \hspace{0.2em} (-log \hspace{0.2em} (\frac{P_i}{\hspace{0.2em} t-1 \hspace{0.2em}}) \times \frac{I_i}{\hspace{0.2em} \sum_{j=1}^{k} I_j \hspace{0.2em}})
    \end{aligned}
\end{equation}
where $P_i$ and $I_i$ respectively represents the $i$-th element in the $\textit{P}_{\textit{conf}}$ and $\mathcal{I}$, and $k$, $t$ are the hyper-parameters defined previously. An important characteristic to be noticed is that \textit{MPCS} approximates \textit{CE} in the limit of $k=1$ and $t \to \infty$. In other words, \textit{CE} can be viewed as a special case of \textit{MPCS} under the limit condition. We illustrate the whole process of \textit{MPCS} calculation in Algorithm~\ref{alg:algorithm}, including the construction of meta pattern (line 3-5, 8), the calculation of concern degree (line 6-7, 9-17) and interval punishment (line 18-20), and the calculation of the \textit{MPCS} value in the whole dataset (line 21-24).

\section{Evaluation and Discussion}
In this section, we first evaluate \textit{MPCS} in general effectiveness and efficiency, and then illustrate its specific advantages in introducing the human values by a case study. Finally, we also discuss the contribution of \textit{MPCS} based on the experiment results. The code and detailed experiment
records are available on our Github repository: \textbf{https://github.com/FlaAI/MPCS}.

\subsection{Experimental Setup}\label{subsec:setup}
There are six datasets from four sources used in the experiments, namely the IRIS and DIGITS from scikit-learn \cite{scikit-learn}, the MNIST, the pedestrian counting dataset CT and the smooth subspace clustering dataset SS from UCR time-series archive \cite{dau2019ucr}, and the synthetic Autonomous Driving Dataset (ADD) generated by the Scenic scenario programming language and the Carla simulator \cite{zhang2021meta}. Different models, including Multi-Layer Perceptron (MLP), Convolutional Neural Network (CNN), Recurrent Neural Network (RNN) and Long Short-Term Memory (LSTM), are adopted and \textit{Adam} is chosen as the optimizer uniformly. The \textit{Accuracy}, $F_{1} \text{-}$\textit{score}, \textit{MCC}, \textit{MS} and \textit{CE} mentioned in section \ref{subsec:basic} are the benchmarks adopted. The experiments are implemented with Python 3.8.8, and PyTorch 1.10.1. 

\begin{figure*}[htb]
\centering
\subfigure
{
    \begin{minipage}[b]{.45\linewidth}
        \centering
        \includegraphics[scale=0.57]{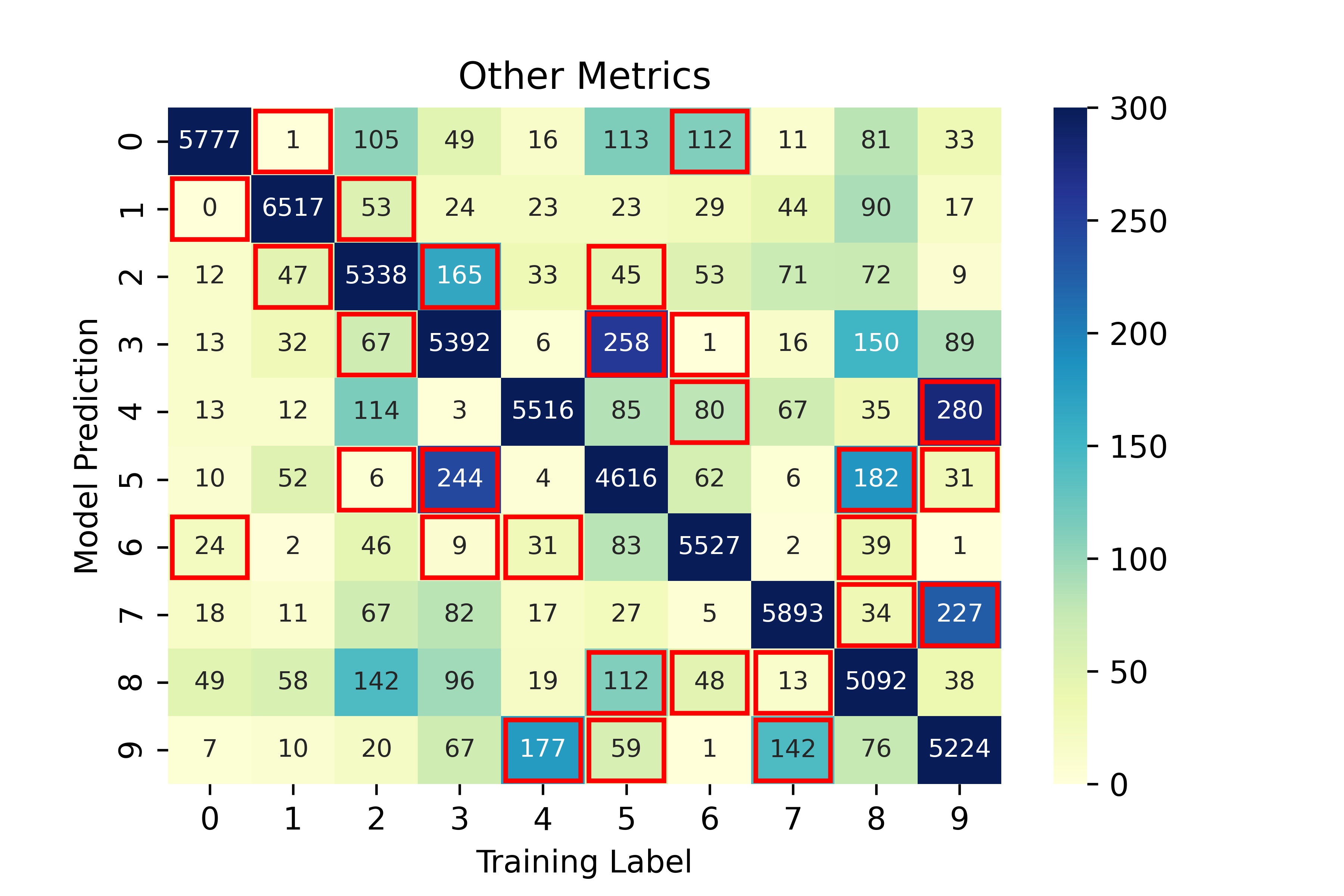}
    \end{minipage}
}
{
    \begin{minipage}[b]{.45\linewidth}
        \centering
        \includegraphics[scale=0.57]{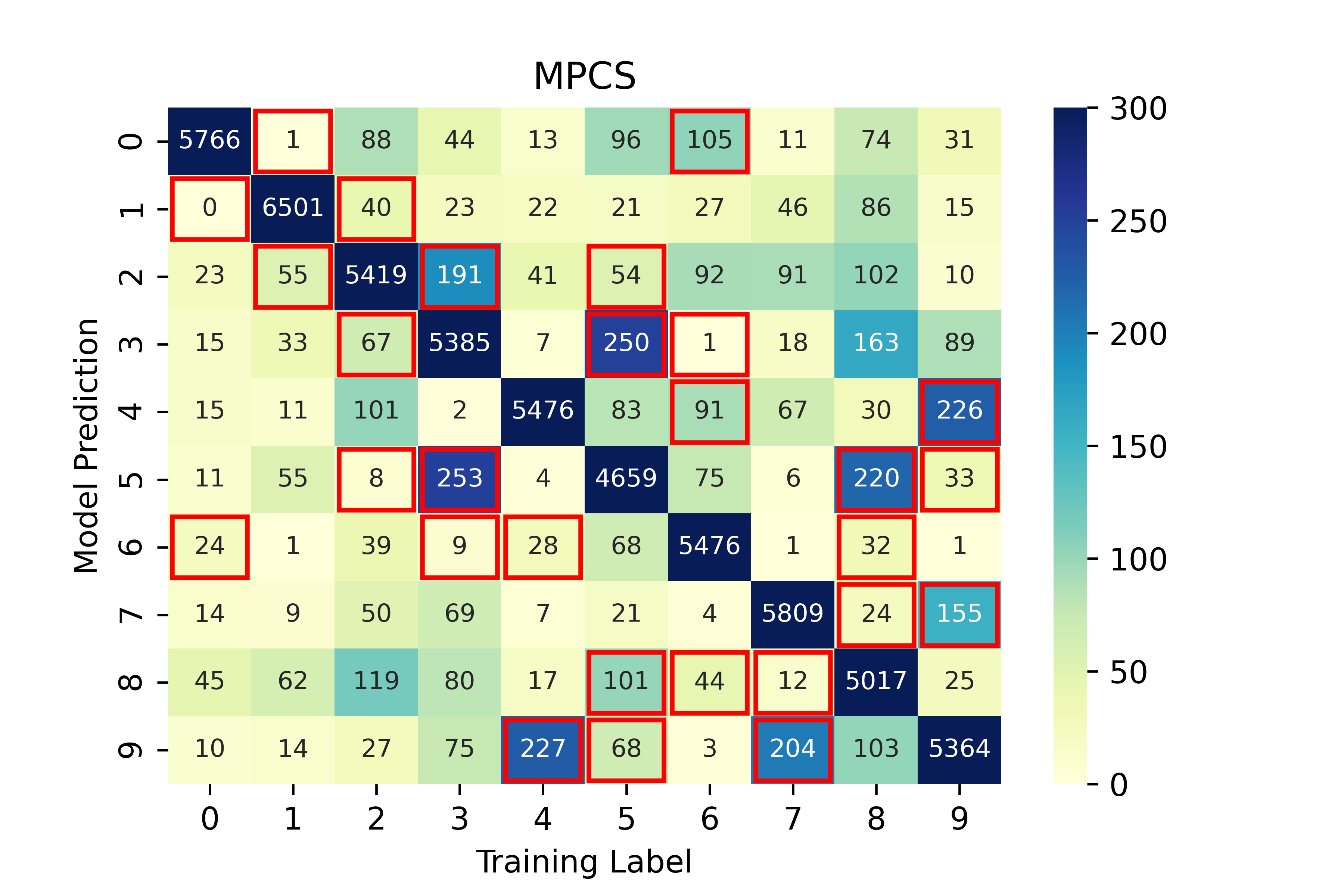}
    \end{minipage}
}
\caption{The figures show the confusion matrices of the prediction results of the two multi-classifiers respectively picked out by the benchmark metrics (they all pick the same model in this cases) and \textit{MPCS} given specific human values (the wrong predictions framed in red are less destructive according to the t-SNE \cite{van2008visualizing}). The model picked by \textit{MPCS} reduces 0.53\% of dangerous cases by only sacrificing 0.04\% of training accuracy.}
\label{fig:Compare}
\end{figure*}

\subsection{General Effectiveness and Efficiency}\label{subsec:effs}
In this section, we evaluate \textit{MPCS} in two general aspects. For one thing, through the comparison of the value and trend between \textit{MPCS} and the benchmark metrics, we verify its effectiveness as an  evaluation measure in the training process of multi-classifier. For another, by illustrating the time cost of \textit{MPCS} calculation, we clarify that it has considerable efficiency as the benchmark metrics. 

\subsubsection{Effectiveness}
Inspired by the evaluation method adopted in \cite{baluta2021scalable}, we use the \textit{spearman correlation coefficient} to calculate the similarity value between \textit{MPCS} and the benchmarks, and regard this kind of relationship as the verification of its effectiveness. We illustrate the results in Table~\ref{tab:Similarity} with corresponding figures as a supplement in Fig.~\ref{fig:10}. From the results, we can find that all the similarity values exceed $\pm 0.9$, and more than half of them are even above $\pm 0.97$, based on which we can give an empirical inference that \textit{MPCS} is highly correlated with the benchmark metrics, so it can indeed serve as an evaluation measure with practical significance.

\begin{table}[htbp]
\caption{The computational time costs of \textit{MPCS} and the benchmark metrics are also similar in the experiments in Table~\ref{tab:Similarity}.}
\label{tab:Time}
\begin{center}
\renewcommand{\arraystretch}{1.2}
\resizebox{8.8cm}{!}{
\begin{tabular}{|c|c|c|c|c|c|c|}
\hline
\multirow{2}{*}{\textbf{ID}} & \multicolumn{6}{|c|}{\bfseries Time Cost ($\times 10^{-3}s$)} \\
\cline{2-7} 
& \textbf{ACC} & \textbf{F1} & \textbf{MCC} & \textbf{MS} & \textbf{CE} & \textbf{MPCS} \\
\hline
\bfseries Exp. 1 & 0.28 & 1.76 & 2.01 & 0.49 & 0.49 & 4.27\\
\hline
\bfseries Exp. 2 & 0.33 & 1.62 & 2.79 & 0.59 & 0.49 & 13.66\\
\hline
\bfseries Exp. 3 & 26.15 & 50.21 & 88.63 & 33.88 & 33.74 & 497.49\\
\hline
\bfseries Exp. 4 & 31.56 & 32.98 & 33.90 & 31.80 & 35.55 & \bfseries43.98\\
\hline
\bfseries Exp. 5 & 692.03 & 696.62 & 702.07 & 693.26 & 788.27 & \bfseries710.92\\
\hline
\bfseries Exp. 6 & 7.25 & 8.34 & 8.17 & 7.26 & 7.59 & \bfseries7.08\\
\hline
\bfseries Exp. 7 & 21.30 & 22.27 & 22.34 & 21.39 & 22.43 & \bfseries22.40\\
\hline
\bfseries Exp. 8 & 38.92 & 40.05 & 39.99 & 39.04 & 40.07 & \bfseries38.89\\
\hline
\bfseries Exp. 9 & 228.87 & 230.86 & 232.93 & 229.05 & 245.07 & \bfseries230.59\\
\hline
\bfseries Exp. 10 & 2672.93 & 2703.11 & 2746.43 & 2685.93 & 2927.46 & \bfseries2802.63\\
\hline
\end{tabular}
}
\end{center}
\end{table}

\subsubsection{Efficiency}
We verify the efficiency of \textit{MPCS} by comparing its computational time cost with the benchmark metrics. Specifically, we calculate them in the whole training dataset for every turn of the training process with a total of 150 turns and take the average time cost as the final results. As illustrated in Table~\ref{tab:Time}, it can be found that the time costs of all the metrics are about the same order of magnitude in general, which is especially clear in relatively complex models.

\subsection{Specific Advantages given Human Values}
So far we have confirmed that \textit{MPCS} can work as a general evaluation measure, but the particular benefits it can bring given specific human values have not been discussed. In this section, we use a case study to illustrate how to use \textit{MPCS} to introduce human values into the evaluation and even the learning of the multi-classifier. We build the case with MLP and MNIST like Exp. 3, and determine the less destructive misclassifications according to the t-SNE projection \cite{van2008visualizing}, a technique that can plot 2D embeddings for high-dimensional datasets while keeping the distance between the samples the same. To be specific, if the incorrect prediction result of a sample is adjacent to its correct label in the projected clusters, this mistake is less destructive compared to confusing non-adjacent labels. So we pick all that kinds of mistakes, which are framed in red in Fig.~\ref{fig:Compare}, into the release list $\mathcal{R}$. Notice that here we just take this as an example since its meaning is relatively simple to understand, while actually the release condition can be customized according to any specific complex requirements in practice.

\subsubsection{\textit{MPCS} for Evaluation}
We train MLP using \textit{CE} for 150 epochs and record the model parameters from each epoch as different candidates, then respectively use different metrics to pick out the optimal models among them. The $f_{\mathcal{R}}$, $k$ and $t$ of \textit{MPCS} is set to be 0.5, 5 and 200 here. In this case, all the metrics except \textit{MPCS} pick out the same model, and the prediction results of the two models are respectively shown in Fig.~\ref{fig:Compare} as the form of the confusion matrix. As can be calculated from the two confusion matrices, although the total number of misclassification of the model picked by the \textit{MPCS} (5128) is greater than that of the model picked by other metrics (5108), the number of destructive cases of the former (2605) is less than that of the latter (2621). In other words, we pick out a model that trades off training accuracy from $91.49\%$ to $91.45\%$ for a reduced destructive rate from $51.31\%$ to $50.78\%$ by applying \textit{MPCS}. This result can especially make sense in real-world safety-critical applications.

\begin{figure*}[htb]
\centering
\subfigure
{
    \begin{minipage}[b]{.22\linewidth}
        \centering
        \includegraphics[scale=0.31]{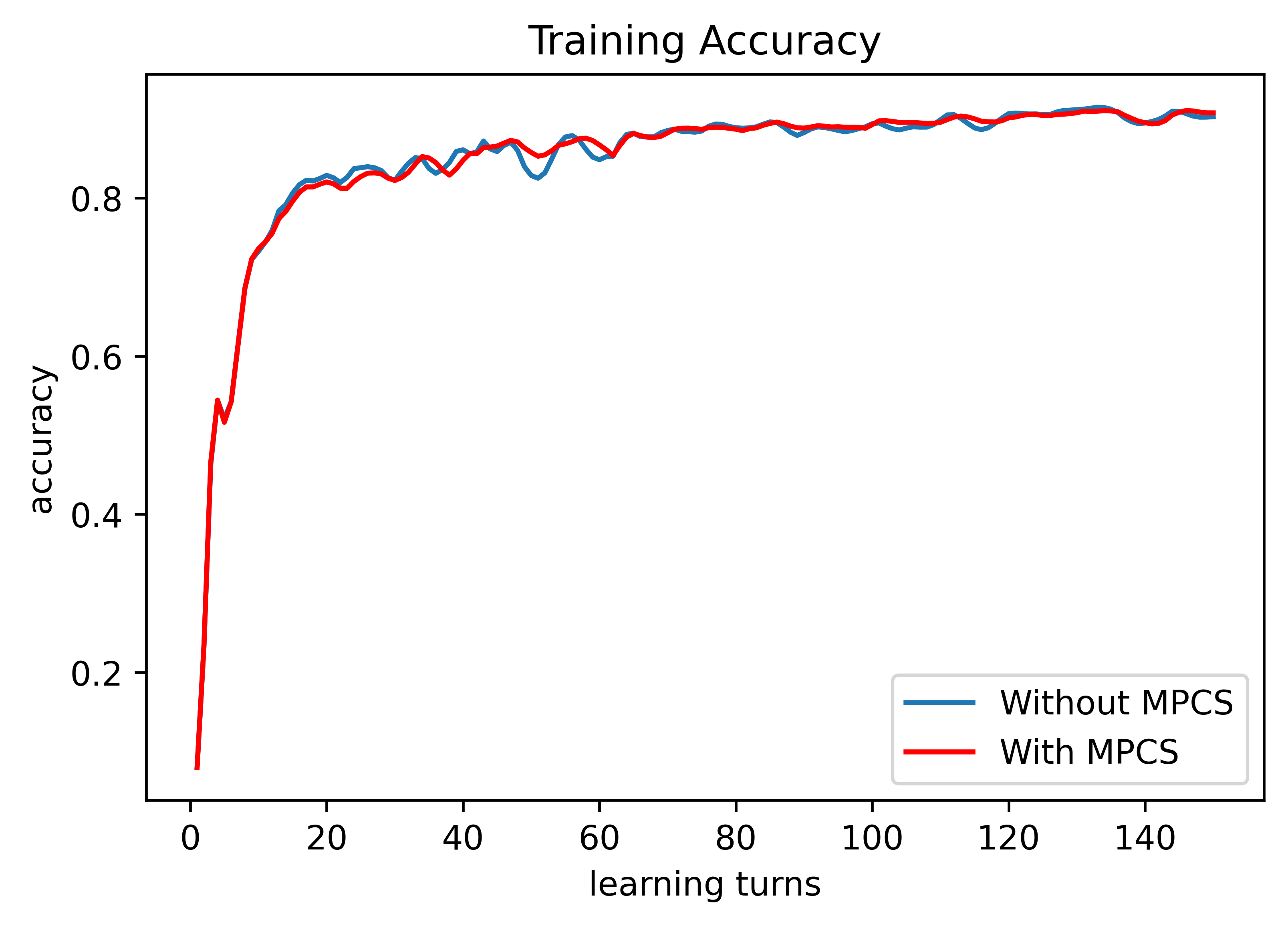}
    \end{minipage}
}
{
    \begin{minipage}[b]{.22\linewidth}
        \centering
        \includegraphics[scale=0.31]{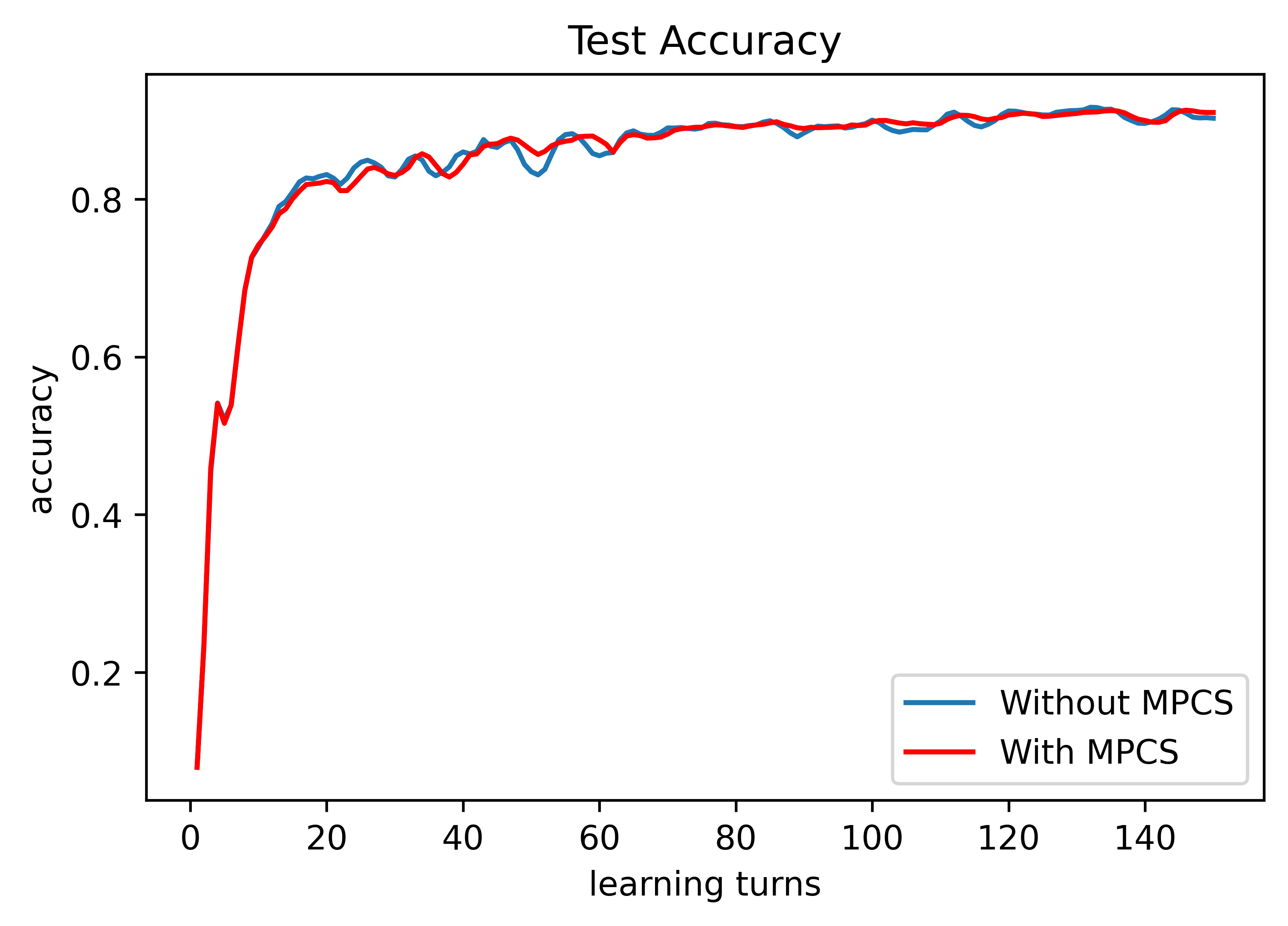}
    \end{minipage}
}
{
    \begin{minipage}[b]{.22\linewidth}
        \centering
        \includegraphics[scale=0.31]{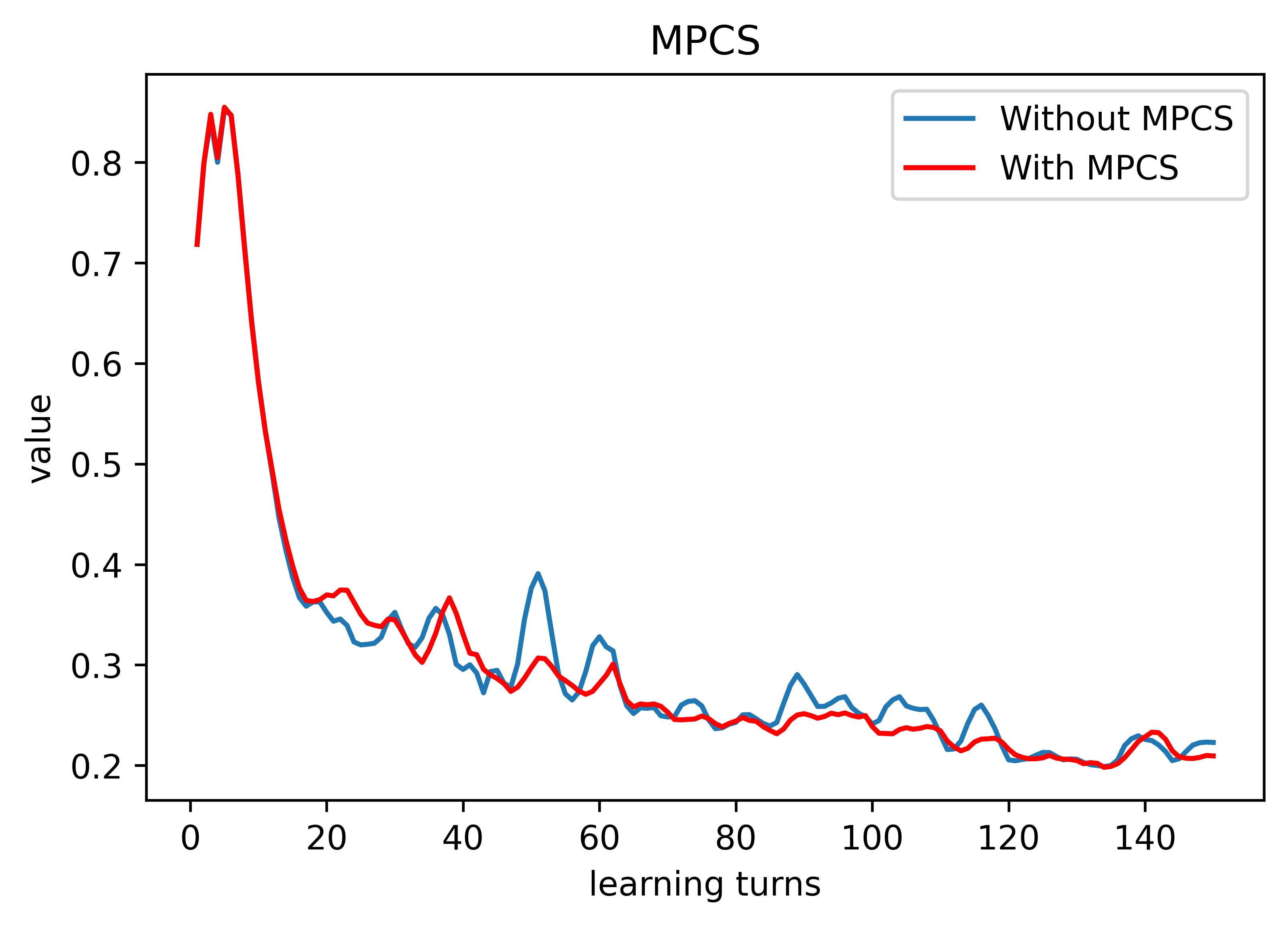}
    \end{minipage}
}
{
    \begin{minipage}[b]{.22\linewidth}
        \centering
        \includegraphics[scale=0.31]{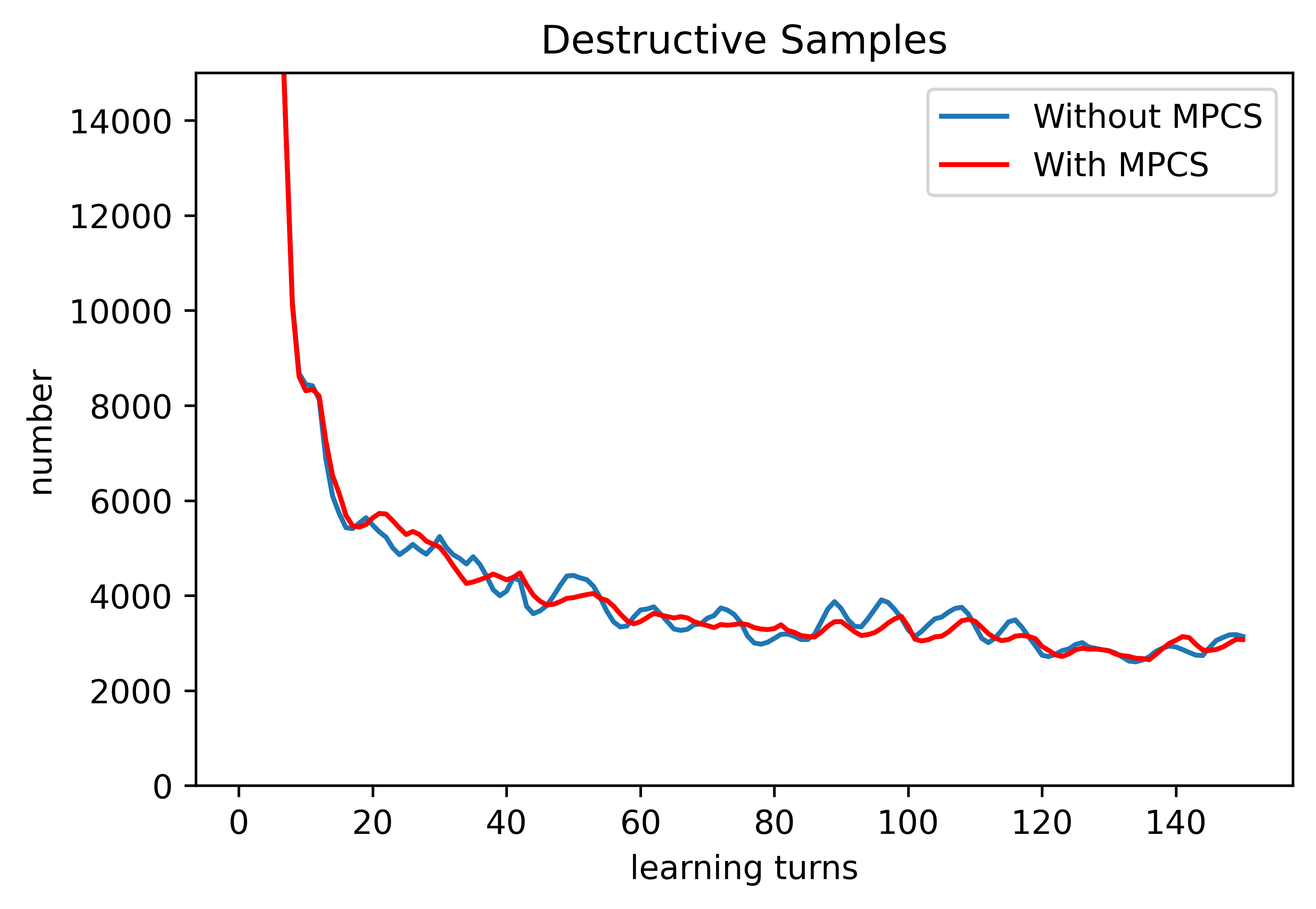}
    \end{minipage}
}
\caption{The figures show the trend of training and test \textit{Accuracy}, the \textit{MPCS} value and the number of dangerous samples among the learning of a multi-classifier respectively with and without \textit{MPCS} given the same human values as Fig.~\ref{fig:Compare} introduced to dynamically adjust the learning rate.}
\label{fig:4}
\end{figure*}

\subsubsection{\textit{MPCS} for Learning}
In addition to the evaluation, we also explore how to introduce the human values into the learning by \textit{MPCS}. Since the abstract representation makes \textit{MPCS} non-differentiable, which means it can not be directly used as a loss, we can just introduce it to refine the training process by dynamically adjusting the learning rate according to its value. This is because the loss is basically a punishment to be learned, so for a smaller (i.e. better) \textit{MPCS} value gotten, we can reduce the punishment by a proportionately smaller learning rate of the current turn. In Fig.~\ref{fig:4}, we record the change of the \textit{Accuracy} in training and test set, the \textit{MPCS} value and the number of destructive samples among the model learning with and without \textit{MPCS} respectively. With \textit{MPCS} introduced, the newly trained model averagely outperforms the original one with a 1.62\% lower \textit{MPCS} value and 0.36\% fewer number of destructive samples. At the same time, there are even slight improvements under the \textit{Accuracy} measure as well, namely 0.04\% and 0.03\% for training and test data.

\subsection{Contributions}
With all the results shown above, now we are ready to answer what can \textit{MPCS} contribute to the community. For the first time, we provide a general idea to introduce two kinds of specific human values into the evaluation and even Learning of multi-classifiers. Different from common metrics having a fixed form all the time, the \textit{MPCS} allows people to flexibly declare what they care more about the model in different practices, and try to cater to their specific will to pick out the optimal model, under the premise of not violating the common metrics too much and having a similar time cost as them. 

On the other hand, technically speaking, \textit{MPCS} is designed regarding the advantages of the existing two common kinds of metrics. Specifically, it provides an abstract view of model output to enable customization as the confusion matrix-based evaluation measures, and calculates its value in a way that approximates \textit{CE} in the limit to make the picked model approximate the posterior probability distribution as much as possible like the loss. And for the negative labels, \textit{MPCS} neither ignores them completely like \textit{CE}, nor gives them the same importance as \textit{MS}, while it adopts a compromise to avoid their shortcomings and makes the assessment more reasonable. Although \textit{MPCS} can still not avoid discrete values in non-limit conditions, not only the granularity is adjustable, but also the extent of discreteness is much lower than the confusion matrix-based measures in general.

\section{Conclusion}
In this paper, to acquire better multi-classifiers given specific human values, we proposed a novel evaluation measure called Meta Pattern Concern Score. It not only achieves considerable effectiveness and efficiency as common metrics in general tasks, but also shows particular advantages under given human values. The \textit{MPCS} is expected to support the customized evaluation and even training of multi-classifiers in real-world practice, especially in safety-critical areas with various human values to be considered. In the future, we plan to evaluate and refine several existing applications by using \textit{MPCS} customized for them, to make them safer and more trustworthy for the public.

\end{document}